\documentclass{article} 
\usepackage{iclr2027_conference,times}

\usepackage{amsmath,amsfonts,bm}

\def\eqref#1{equation~\ref{#1}}

\def\1{\bm{1}}

\DeclareMathAlphabet{\mathsfit}{\encodingdefault}{\sfdefault}{m}{sl}
\SetMathAlphabet{\mathsfit}{bold}{\encodingdefault}{\sfdefault}{bx}{n}

\usepackage{hyperref}
\usepackage{url}
\usepackage{xcolor}
\usepackage{graphicx}
\usepackage{booktabs}
\usepackage{wrapfig}
\usepackage{caption}
\usepackage{subcaption}
\usepackage{amsmath,amssymb}
\usepackage{booktabs,array,longtable}
\usepackage[table]{xcolor}
\definecolor{metricgreen}{RGB}{52,180,20}
\newcommand{\gcell}[2]{\cellcolor{metricgreen!#1}#2}
\newcommand{\pcell}[1]{\cellcolor{gray!12}#1}

\title{Enhancing Autoregressive Video Generation \\ via Representation Adversarial Distillation}

\author{%
  Fangyu Lin\textsuperscript{1,2}, 
  Xingtong Ge\textsuperscript{1,2},
  Lunjie Zhu\textsuperscript{1,2},
  Yi Zhang\textsuperscript{2\textsuperscript{$\ddagger$}}, 
  Zhening Liu\textsuperscript{1},
  Tianhang Wang\textsuperscript{3},\\
  \textbf{Mengfei Li\textsuperscript{1,2},
  Yumeng Zhang\textsuperscript{1},
  Guanglu Song\textsuperscript{2},
  Yu Liu\textsuperscript{2},
  Jun Zhang\textsuperscript{1}\textsuperscript{$\dagger$}}  \\
  $^1$The Hong Kong University of Science and Technology, \\
  $^2$Vivix Group Limited,
  $^3$Zhejiang University \\
  \texttt{rslinfy@gmail.com, eejzhang@ust.hk}
}

\iclrfinalcopy 
\begin{document}
\maketitle

\begingroup
\renewcommand{\thefootnote}{\fnsymbol{footnote}}
\setcounter{footnote}{3}
\footnotetext{Project Lead}
\setcounter{footnote}{2}
\footnotetext{Corresponding Author}
\endgroup

\lhead{Preprint}

\begin{abstract}
Few-step autoregressive video generation enables efficient streaming synthesis, but errors introduced in early temporal blocks are reused as context and can propagate through subsequent rollouts, leading to detail degradation, structural drift, and unstable motion. Existing distribution matching distillation (DMD) primarily aligns student and teacher distributions in diffusion latent space, but provides no direct supervision over the perceptual quality of decoded videos. We introduce \textbf{Radian}, a representation-space adversarial distillation framework that complements on-policy DMD with real-data adversarial supervision in the feature space defined by a frozen visual foundation model (VFM). During training, Radian sparsely decodes frames from autoregressive student rollouts, extracts multi-level visual representations, and applies lightweight discriminator heads to distinguish generated outputs from real video frames. The DMD objective anchors the student to the pretrained teacher, while the representation-space adversarial objective supplies complementary perceptual and semantic gradients that promote high-quality modes. These additional components are discarded after training, leaving the generator architecture and inference-time denoising budget unchanged. Experiments on Wan2.1-1.3B cover four-step chunk-wise, one-step frame-wise, and minute-long autoregressive generation. Our method achieves a VBench Total of $0.8444$ and a VideoAlign Total of $0.8033$ under four-step generation, and improves VBench-Long from $0.7805$ to $0.8041$ over Rolling Forcing while using fewer denoising steps. Controlled comparisons across image, video, and diffusion representations further indicate that the choice of representation spaces induces distinct adversarial signals, and external VFM gradients complement DMD more effectively than adversarial supervision derived from diffusion-internal features. Project page: \href{https://rslinfy.github.io/Radian-Project-Page}{https://rslinfy.github.io/Radian-Project-Page}.
\end{abstract}

\section{Introduction}
\label{sec:introduction}

Recent video generation models have substantially advanced visual fidelity, motion realism, and prompt alignment~\citep{wan,veo3,hunyuanvideo,seedance,ltx2}. However, typical video generators synthesize fixed-length clips through bidirectional temporal attention, incurring prohibitive latency, computational costs and incompatibility with causal streaming generation. These limitations conflict with today's urgent demand for interactive world models, real-time audio-visual avatars, 3D telepresence, and online video editing, which require continuous generation and immediate response~\citep{minwm,worldplay,liveavatar,ola,rosmi}. The central challenge is to retain the quality of a bidirectional teacher under a small sampling budget and long autoregressive rollouts.

Most previous works address the first challenge through few-step diffusion distillation. Progressive and consistency-based methods compress iterative denoising by direct mappings between noise levels~\citep{progressive_distillation,consistency_models,lcm,scm,rcm}, while Distribution Matching Distillation (DMD) trains a student to reproduce the output distribution of a pretrained diffusion teacher~\citep{dmd,dmd2}. Recent video generation methods like Self Forcing further combine distillation with block-wise causal generation and KV caching, and train the student on its own generated histories to reduce the train-test gap~\citep{causvid,self_forcing,causal_forcing,anyflow,salt,rolling_forcing,self_forcing_pp}. Nevertheless, early generated blocks with errors become the context for subsequent blocks, and these errors then propagate and accumulate over generation, leading to visual degradation, detail loss, and semantic drift. Moreover, methods like DMD transfer the teacher distribution through score differences in the diffusion latent space, failing to align decoded outputs with real videos.


\begin{figure*}[t]
    \centering
    \noindent
    \makebox[\textwidth][c]{%
        \includegraphics[width=.32\textwidth]{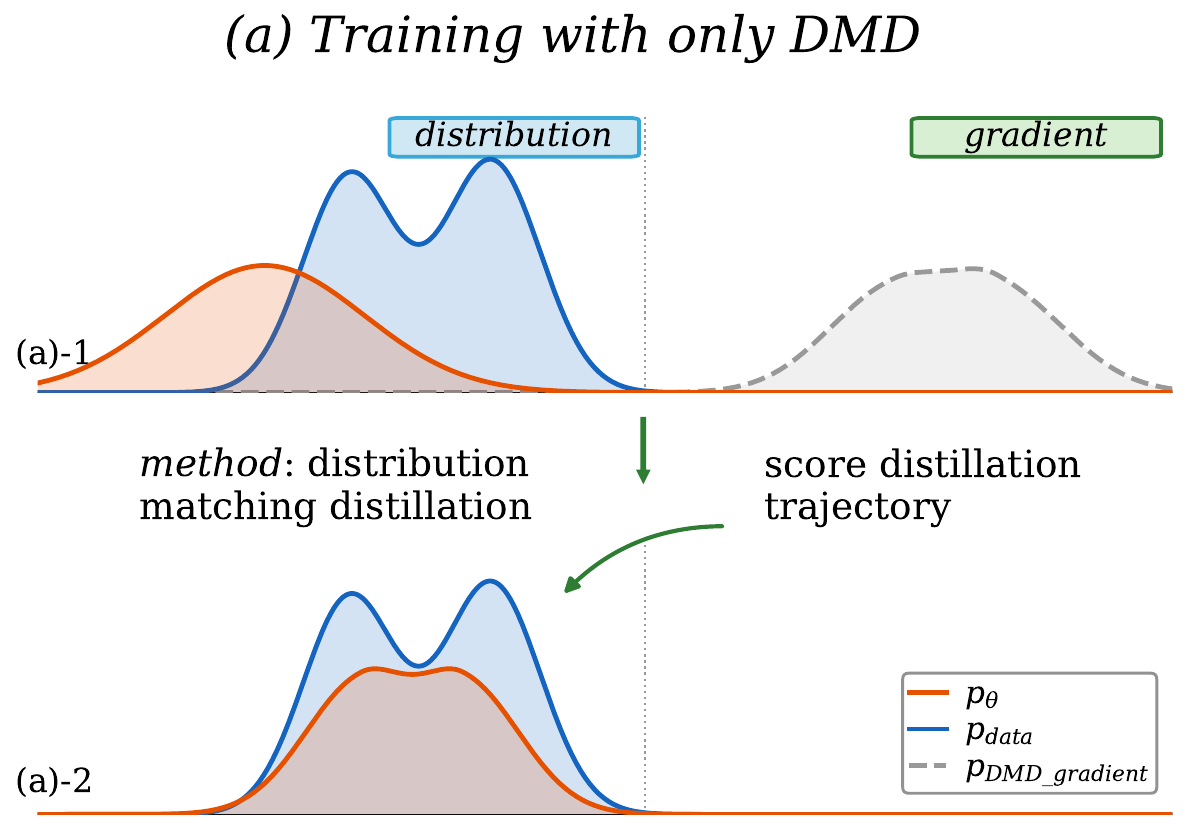}%
        \includegraphics[width=.32\textwidth]{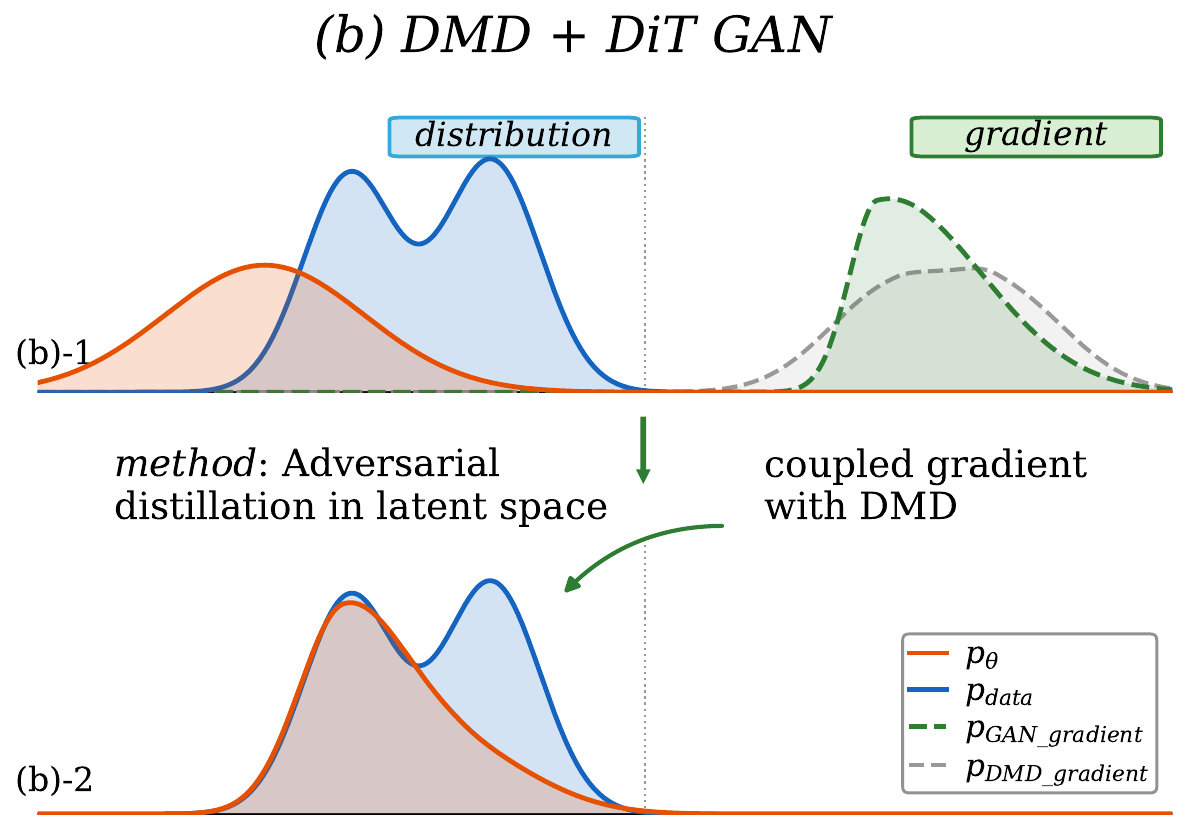}%
        \includegraphics[width=.344\textwidth]{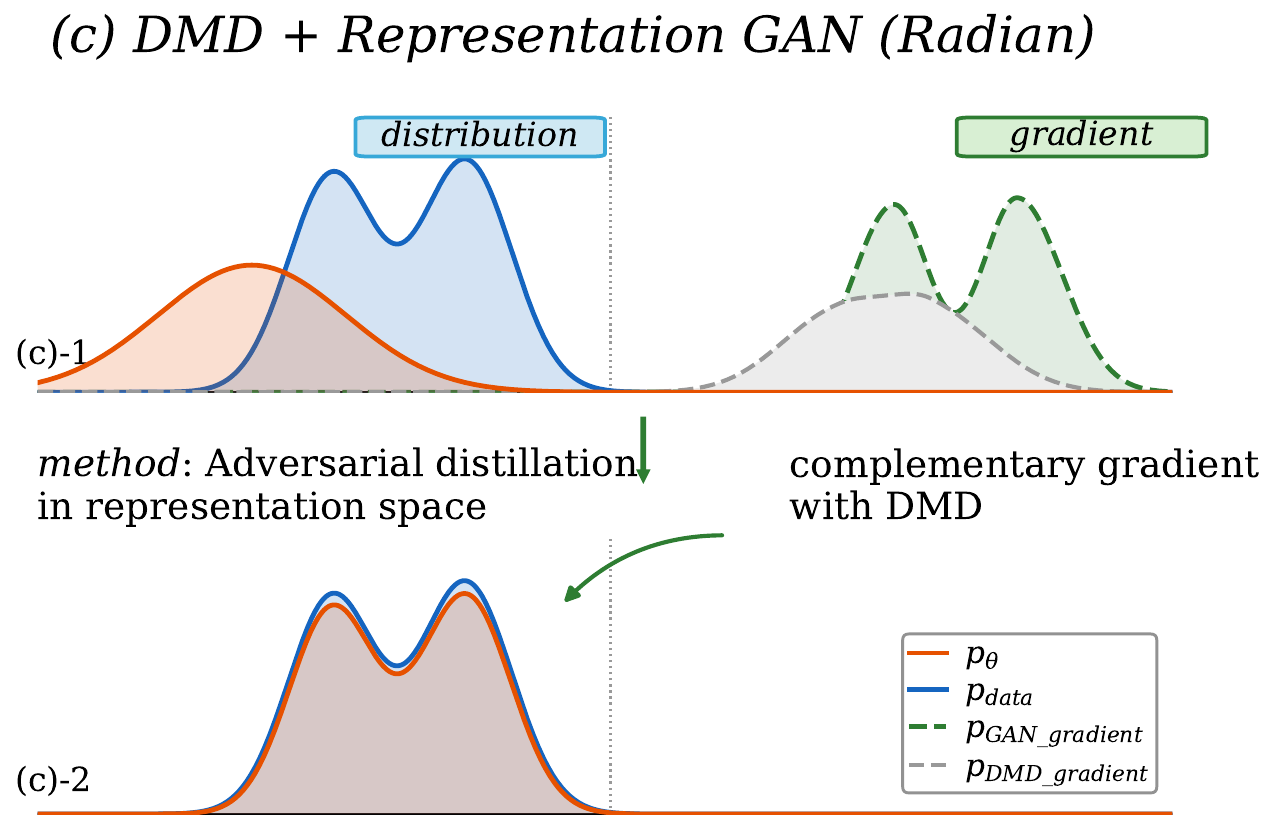}%
    }
    \caption{\textbf{Complementary Representation-Space Supervision.}
    \textbf{Radian} complements DMD with perceptual gradients, guiding the student toward a better-matched data distribution.}
    \label{fig:teaser}
    \vspace{-1.4em}
\end{figure*}

Another line of works improve few-step generators through adversarial distillation~\citep{LADD,APT,AAPT,one_forcing,AAD1,sdxl_lightning,VPAE}. When real samples are used as the positive reference, the discriminator provides a measure of data realism that teacher-based score matching does not offer. ADD exploits this principle for image generation in the representation-space of a frozen DINOv2 encoder~\citep{ADD}. When it comes to video generation, in contrast, discrimination is performed in clean or noise-corrupted VAE latents or internal features from DiT backbones. However, such discrimination may be sub-optimal for improving the video quality, as VAE latents are optimized for compression and reconstruction, whereas diffusion-backbone features are learned for denoising or velocity prediction. The diffusion space used by DMD is close to them, causing adversarial gradients to overlap with score distillation. Moreover, without RGB decoding, the discriminator cannot detect texture blur, artifacts, or VAE decoding errors, making such features suboptimal for adversarial supervision.

In this work, we present \textbf{Radian}, a method that \textbf{performs adversarial distillation in representation space} to enhance few-step autoregressive video generation. Radian augments on-policy DMD with an adversarial objective over features from a frozen visual foundation model (VFM). During training, selected outputs from causal autoregressive rollouts are decoded into RGB frames and mapped to multi-level VFM features, which are distinguished from those of real videos using lightweight discriminator heads. The VFM and discriminator heads are removed after training, leaving the generator architecture and inference cost unchanged. DMD anchors the student to the teacher distribution, while representation-space adversarial supervision uses real data to suppress perceptually and semantically degraded outputs and reshape the learned distribution toward higher-quality modes.

Our contributions are as follows. We introduce a new representation adversarial distillation framework that complements on-policy DMD with real-data supervision over decoded outputs while preserving the original inference architecture and cost. Beyond the proposed framework, we systematically study adversarial representation spaces spanning image VFMs including DINOv2~\citep{dinov2}, DINOv3~\citep{dinov3}, and SigLIP2~\citep{siglip2}, video encoders including V-JEPA 2.1~\citep{vjepa21} and VideoMAE~\citep{videomae}, as well as diffusion-internal features. Our analysis further shows that different image representations induce distinct adversarial signals, video representations provide complementary temporal regularization, and external VFM gradients are more complementary to DMD than diffusion-internal supervision as conceptually illustrated in Fig.~\ref{fig:teaser}. Finally, experiments on causal generators built upon Wan2.1 1.3B demonstrate consistent improvements in visual quality, motion quality, and prompt alignment over strong Forcing-series baselines, achieving state-of-the-art performance on VBench~\citep{huang2024vbench} and VBench-Long~\citep{huang2026vbenchpp} across 4-step, 1-step, and long-horizon generation.

\section{Related Work}
\label{sec:related_work}


\textbf{Few-Step and Autoregressive Video Generation.}
Few-step diffusion distillation generally follows trajectory- or distribution-based paradigms. 
Trajectory-based methods compress teacher sampling trajectories through progressive or consistency objectives~\citep{progressive_distillation,consistency_models,lcm,scm,rcm,causal_rcm}, whereas distribution-based methods directly align student and teacher output distributions. 
DMD estimates a reverse Kullback--Leibler (KL) gradient from teacher and student score functions, and DMD2 further improves score estimation and training stability~\citep{dmd,dmd2,senseflow}. 
Related variants further adapt distribution matching to efficient few-step video generation~\citep{anyflow,salt}.
For streaming generation, autoregressive video models factorize videos into causal temporal blocks and reuse previous outputs through KV caching. 
CausVid~\citep{causvid} distills bidirectional video diffusion into a few-step causal generator, while Self Forcing~\citep{self_forcing} reduces the training--inference gap by training on the student's own rollouts. 
Causal Forcing and Causal Forcing++ further improve initialization and scalable few-step frame-wise generation~\citep{causal_forcing,causal_forcing_pp}, while Rolling Forcing and Self Forcing++ extend self-rollout training to long-horizon generation~\citep{rolling_forcing,self_forcing_pp}.
Recent works further explore complementary mechanisms for long-horizon streaming generation: Salt++ aligns causal contexts across generator sampling and score estimation for few-step multimodal generation \citep{saltpp}, while Spatia maintains an updatable 3D memory to improve long-term spatial consistency \citep{spatia}.


\textbf{Adversarial Training for Diffusion Models.}
Adversarial objectives have been widely used to preserve sample quality under small inference budgets. In image generation, ADD combines diffusion score distillation with a discriminator on frozen DINOv2 features~\citep{ADD}, while related methods also exploit internal diffusion representations~\citep{LADD}. Extending to video generation, APT applies adversarial post-training to one-step synthesis~\citep{APT}, and AAPT further adapts it to autoregressive video generation~\citep{AAPT}. Subsequent work explores adversarial distribution matching, cross-step self-distillation, phased training, and other autoregressive adversarial formulations~\citep{ADM,ASD,VPAE,AAD1,one_forcing,adversarial_flow_distillation}. Nevertheless, most existing methods discriminate with diffusion-internal features, which may provide redundant or conflicting supervision with DMD and remain weakly aligned with decoded videos, limiting sensitivity to pixel-space artifacts.

\textbf{Pretrained Representations for Generative Modeling.}
Pretrained representation spaces are increasingly used in generative modeling~\citep{dinov2,dinov3,ijepa,vjepa21,videomae,decq}. REPA~\citep{repa} aligns intermediate DiT states with clean VFM features, while RAE~\citep{rae} adopts frozen VFM encoders as generative latent spaces. Subsequent works further explore structured representations and representation-space generative supervision~\citep{svg,decq,perceptual_flow_matching,representation_distribution_matching}. In contrast, our proposed method targets an already distilled autoregressive generator and adversarially distinguishes decoded real and generated videos using frozen VFM features, rather than aligning features or redesigning the latent space. This provides real-data supervision complementary to DMD without additional inference cost.



\section{Method}
\label{sec:method}

\subsection{Preliminaries}
\label{sec:preliminaries}

\noindent\textbf{Flow-Matching Models and Causal Video Generation.} Let $\mathbf{x}_0$ denote a clean video latent and $\mathbf{x}_t=\alpha_t\mathbf{x}_0+\sigma_t\boldsymbol{\epsilon}$ its noisy state at timestep $t$, where $\boldsymbol{\epsilon}\sim\mathcal{N}(\mathbf{0},\mathbf{I})$, and $\alpha_t$ and $\sigma_t$ denote the signal and noise scaling coefficients determined by the diffusion noise schedule, respectively. A conditional diffusion or flow-matching model learns a time-dependent vector field, $\mathbf{v}(\mathbf{x}_t,t,c)$, that transports noise toward the data distribution through the corresponding probability-flow ODE. For causal generation, the video is partitioned into $B$ temporal blocks, $\mathbf{x}_0=(\mathbf{x}_0^{1},\ldots,\mathbf{x}_0^{B})$, and its distribution is factorized as
\begin{equation}
p_{\theta}(\mathbf{x}_0\mid c)=\prod_{b=1}^{B}p_{\theta}\!\left(\mathbf{x}_0^{b}\mid\mathbf{x}_0^{<b},c\right).
\end{equation}
Each block is generated with a few denoising steps while previously computed context is reused through the KV cache. During on-policy training, the student conditions on its own generated history $\hat{\mathbf{x}}_0^{<b}$, matching context at inference time \citep{flowmatching,causvid,self_forcing}.

\noindent\textbf{Distribution Matching Distillation.} DMD trains a few-step generator by minimizing the reverse KL divergence between the student distribution and a reference distribution defined by a pretrained diffusion teacher \citep{dmd,dmd2}. Given an autoregressive student rollout $\hat{\mathbf{x}}_0=G_{\theta}(\mathbf{z},c)$, we obtain $\hat{\mathbf{x}}_t=\alpha_t\hat{\mathbf{x}}_0+\sigma_t\boldsymbol{\epsilon}$ through forward diffusion. At a randomly sampled timestep $t$, gradient of the reverse KL divergence is approximated by the teacher-student score functions. Specifically, let $\mathbf{s}_{r}(\hat{\mathbf{x}}_t,t,c)$ denote the frozen teacher score and $\mathbf{s}_{\phi}(\hat{\mathbf{x}}_t,t,c)$ be the score of the current student distribution, estimated by a trainable fake-score model. The generator update is
\begin{equation}
\nabla_{\theta}\mathcal{L}_{\mathrm{DMD}}
=
\mathbb{E}_{t,\mathbf{z},\boldsymbol{\epsilon}}
\left[
\left(\mathbf{s}_{\phi}(\hat{\mathbf{x}}_t,t,c)-\mathbf{s}_{r}(\hat{\mathbf{x}}_t,t,c)\right)
\frac{\partial\hat{\mathbf{x}}_t}{\partial\theta}
\right].
\label{eq:dmd}
\end{equation}
The fake-score model and generator are updated alternately, allowing DMD to match the student rollout distribution to the teacher under a small sampling budget.

\noindent\textbf{Adversarial Distillation.} Adversarial distillation introduces supervision from real data by jointly training a generator $G_{\theta}$ and a discriminator $D_{\psi}$ that distinguishes generated samples from real ones \citep{ADD,LADD,APT,AAPT,AAD1}. Given noise $\mathbf{z}\sim\mathcal{N}(\mathbf{0},\mathbf{I})$ and conditioning $c$, the standard adversarial objective is
\begin{equation}
\min_{\theta}\max_{\psi}\;
\mathbb{E}_{\mathbf{x}\sim p_{\mathrm{data}}}
\left[\log D_{\psi}(\mathbf{x},c)\right]
+
\mathbb{E}_{\mathbf{z}\sim\mathcal{N}(\mathbf{0},\mathbf{I})}
\left[\log\!\left(1-D_{\psi}(G_{\theta}(\mathbf{z},c),c)\right)\right].
\end{equation}
Here, $D_{\psi}$ denotes the complete discriminator mapping, including any input transformation or feature extractor. 
For autoregressive video generation, $G_{\theta}(\mathbf{z},c)$ represents a complete causal rollout. In practice, the logistic objective may be replaced by hinge or related adversarial losses. Unlike teacher-based distribution matching, adversarial distillation directly compares generated samples against real data and therefore provides complementary signals toward the real-data distribution.

\begin{figure*}[!t]
    \centering
    \includegraphics[width=\linewidth]{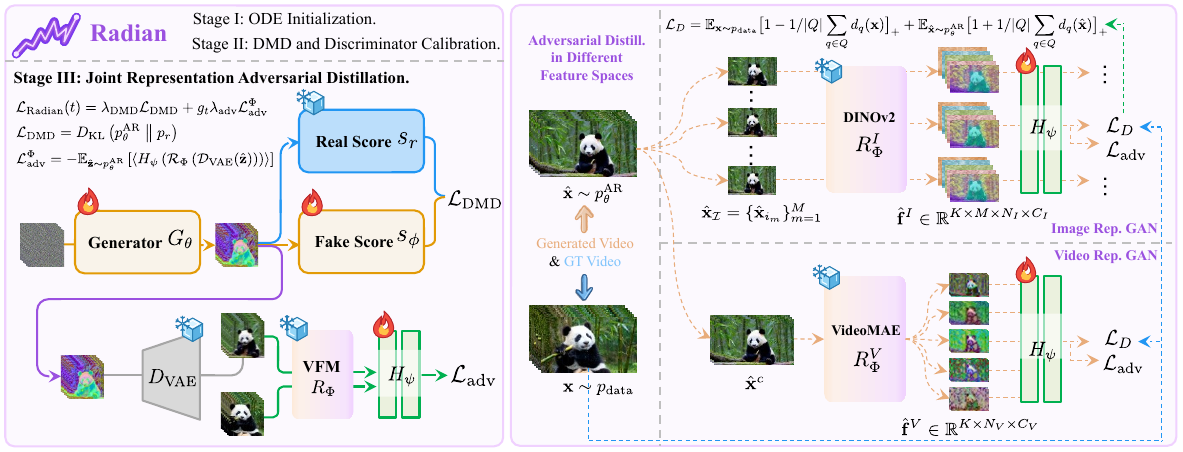}
    \caption{\textbf{Overview of the Three-stage Training Pipeline. }}
    \label{fig:overview}
    \vspace{-4mm}
\end{figure*}

\subsection{Representation Adversarial Distillation for Causal Video Generation}
\label{sec:on_policy_rad}



We consider a causal student generator $G_{\theta}$ that synthesizes a video as a sequence of $B$ consecutive latent blocks, each generated with $K$ denoising steps. We denote the resulting latent rollout by $\hat{\mathbf{z}}=\hat{\mathbf{z}}^{(1:B)}\sim p_{\theta}^{\mathrm{AR}}$. Since the rollout reuses previously computed context through the KV cache, it follows the same causal computation at inference and exposes the student to errors in its own history. We train this causal student through the three successive stages illustrated in Fig.~\ref{fig:overview}.

\noindent\textbf{Stage I: ODE Initialization.} We first initialize the causal student by regressing from intermediate states of precomputed ODE trajectories to their clean endpoints. Given an intermediate latent $\mathbf{z}_{\tau}^{\mathrm{ODE}}$, its corresponding timestep $\tau$, and the clean trajectory endpoint $\mathbf{z}_{0}^{\mathrm{ODE}}$, the initialization objective is
\begin{equation}
\mathcal{L}_{\mathrm{ODE}}
=
\mathbb{E}_{(\mathbf{z}_{\tau}^{\mathrm{ODE}},\mathbf{z}_{0}^{\mathrm{ODE}},\tau,c)}
\left[
\left\|
G_{\theta}
\left(
\mathbf{z}_{\tau}^{\mathrm{ODE}},\tau,c
\right)
-
\mathbf{z}_{0}^{\mathrm{ODE}}
\right\|_{2}^{2}
\right].
\end{equation}
This stage provides the student with a stable few-step causal solver before it is trained on its own autoregressive rollouts, reducing the burden on the subsequent stages.

\noindent\textbf{Stage II: DMD Stabilization and Discriminator Calibration.} Starting from the ODE initialization, we briefly optimize the causal student with on-policy DMD using Eq.~\ref{eq:dmd} to adapt it to its own autoregressive histories. In parallel, the feature discriminator $D_{\psi}$ is warmed up using real videos and decoded student rollouts. A tunable sanity gate blocks adversarial gradients from affecting the generator during this period, preventing an uninitialized discriminator from destabilizing generator training. The resulting student and calibrated discriminator are then used to initialize Stage III.

\noindent\textbf{Stage III: Joint Representation Adversarial Distillation.} Once the sanity gate is activated, the student is jointly optimized by on-policy DMD and representation adversarial distillation:
\begin{equation}
\mathcal{L}_{\mathrm{Radian}}(t)
=
\lambda_{\mathrm{DMD}}\mathcal{L}_{\mathrm{DMD}}
+
g_t\lambda_{\mathrm{adv}}\mathcal{L}_{\mathrm{adv}}^{\Phi}.
\end{equation}

\begin{wrapfigure}{l}{0.4\textwidth} 
  \centering
  \includegraphics[width=\linewidth]{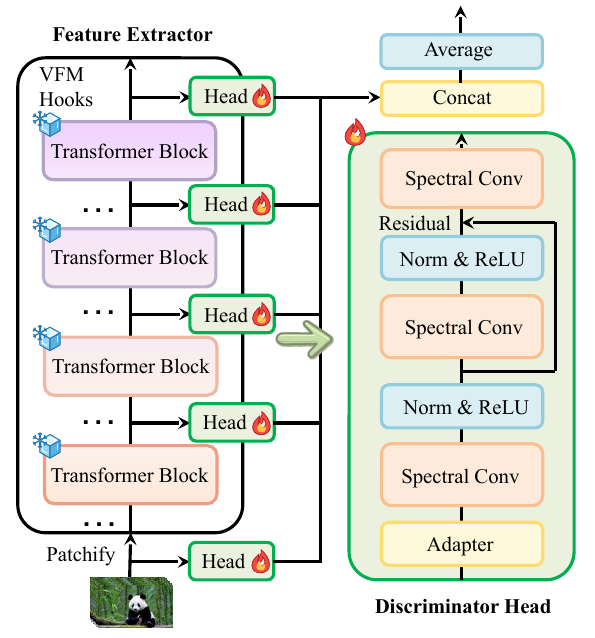} 
  \caption{\textbf{Architecture of Lightweight Discriminator Head.}}
  \label{fig:D}
\end{wrapfigure}

$\mathcal{L}_{\mathrm{DMD}}$ serves as an anchor that keeps the student close to the teacher's distribution, whereas $\mathcal{L}_{\mathrm{adv}}^{\Phi}$ uses real videos to reshape the student distribution toward perceptually realistic and semantically coherent outputs. The two terms are evaluated on separate self-rollouts sampled from the same causal student distribution, with $g_t$ as a weighting coefficient. Both objectives remain active throughout the joint stage, allowing adversarial supervision without discarding the teacher-derived generative prior.

Concretely, to compute the adversarial objective, we define the multi-level feature operator of a frozen visual encoder $\Phi$ as $\mathcal{R}_{\Phi}(\mathbf{x})=\{\Phi_{\ell}(\mathcal{P}(\mathbf{x}))\}_{\ell\in\mathcal{S}}$, with $\mathcal{S}$ being the selected feature levels. The adversarial objective is
\begin{equation}
\mathcal{L}_{\mathrm{adv}}^{\Phi}
=
-
\mathbb{E}_{\hat{\mathbf{z}}\sim p_{\theta}^{\mathrm{AR}}}
\left[
\left\langle
H_{\psi}
\left(
\mathcal{R}_{\Phi}
\left(
\mathcal{D}_{\mathrm{VAE}}(\hat{\mathbf{z}})
\right)
\right)
\right\rangle
\right].
\label{eq:feature_adv}
\end{equation}
Here, $\mathcal{P}$ converts decoded RGB frames into the input format required by $\Phi$ through range conversion, spatial augmentation, resizing, and normalization. $H_{\psi}$ collects the trainable discriminator heads, and $\langle\cdot\rangle$ averages their logits over sampled frames, feature levels, and spatial tokens. $\mathcal{D}_{\mathrm{VAE}}$ represent the VAE decoder. They remain frozen but differentiable during the training process, and are removed after training.

\subsection{Multi-Level Feature Discrimination over Decoded Outputs}
\label{sec:feature_discriminator}

Eq.~\ref{eq:feature_adv} is realized using sparse temporal sampling and multi-level feature discrimination. Given a real video $\mathbf{v}\sim p_{\mathrm{data}}$ and a generated rollout $\hat{\mathbf{z}}\sim p_{\theta}^{\mathrm{AR}}$, we uniformly sample $M$ latent positions without replacement and map them to the corresponding normalized positions in the real video:
\begin{equation}
\begin{aligned}
\mathcal{I}
&=
\{i_m\}_{m=1}^{M}
\sim
\operatorname{UnifWOR}
\left(
\{0,\ldots,T_z-1\},M
\right),
\\
&j_m
=
\operatorname{round}
\left(
\frac{i_m}{T_z-1}(T_r-1)
\right).
\end{aligned}
\end{equation}
where $T_z$ and $T_r$ denote the lengths of the latent rollout and real video. $\operatorname{UnifWOR}$ denotes uniform sampling without replacement, and $j_m$ is the real-video position obtained by linearly mapping the sampled latent index $i_m$ to the normalized temporal coordinate of the real video. For each sampled position, only a short latent neighborhood $\mathcal{N}_{q}(i_m)={i:\lvert i-i_m\rvert\le q}$ of radius $q$ is decoded:
\begin{equation}
\hat{\mathbf{x}}_{m}
=
\left[
\mathcal{D}_{\mathrm{VAE}}
\left(
\hat{\mathbf{z}}_{\mathcal{N}_{q}(i_m)}
\right)
\right]_{\mathrm{ctr}},
\qquad
\mathbf{x}_{m}
=
\mathbf{v}_{j_m}.
\end{equation}
where $[\cdot]_{\mathrm{ctr}}$ selects the center RGB frame at each sampled position. The real and generated videos are sampled independently, and the mapping only aligns their relative temporal positions. For $\mathbf{y}_m\in{\mathbf{x}_m,\hat{\mathbf{x}}_m}$, the frozen encoder extracts multi-level spatial features.
\begin{equation}
\mathbf{F}_{\ell}(\mathbf{y}_m)
=
\Phi_{\ell}\!\left(\mathcal{P}(\mathbf{y}_m)\right)
\in
\mathbb{R}^{N_{\ell}\times C_{\ell}},
\qquad
\widetilde{\mathbf{F}}_{\ell}(\mathbf{y}_m)
=
\mathbf{F}_{\ell}(\mathbf{y}_m)
+
\mathbf{1}_{N_{\ell}}\mathbf{r}_{\ell}^{\top},
\end{equation}
where $N_{\ell}$ and $C_{\ell}$ denote the number and dimension of spatial tokens, and $\mathbf{r}_{\ell}$ is the global readout token. Injecting $\mathbf{r}_{\ell}$ into spatial tokens provides each local feature with global semantic context. As illustrated in Fig.~\ref{fig:D}, on top of these features, each selected level is equipped with a lightweight trainable head $h_{\psi,\ell}$ that produces dense spatial logits:
\begin{equation}
\mathbf{d}_{m,\ell}(\mathbf{y}_m)
=
h_{\psi,\ell}
\left(
\widetilde{\mathbf{F}}_{\ell}(\mathbf{y}_m)
\right)
\in
\mathbb{R}^{N_{\ell}},
\qquad
d_{m,\ell,n}(\mathbf{y}_m)
=
\left[
\mathbf{d}_{m,\ell}(\mathbf{y}_m)
\right]_{n}.
\end{equation}

Finally, let $\mathcal{Q}=\{(m,\ell,n)\mid m\in\{1,\ldots,M\},\,\ell\in\mathcal{S},\,n\in\{1,\ldots,N_{\ell}\}\}$ collect all supervised temporal, feature-level, and spatial locations. The discriminator is trained with hinge loss:
\begin{equation}
\begin{aligned}
\mathcal{L}_{D}
&={}
\mathbb{E}_{\mathbf{v}\sim p_{\mathrm{data}},\,\mathcal{I}}
\left[
\frac{1}{|\mathcal{Q}|}
\sum_{(m,\ell,n)\in\mathcal{Q}}
\left(1-d_{m,\ell,n}(\mathbf{x}_{m})\right)_{+}
\right]
\\
&+
\mathbb{E}_{\hat{\mathbf{z}}\sim p_{\theta}^{\mathrm{AR}},\,\mathcal{I}}
\left[
\frac{1}{|\mathcal{Q}|}
\sum_{(m,\ell,n)\in\mathcal{Q}}
\left(1+d_{m,\ell,n}(\hat{\mathbf{x}}_{m})\right)_{+}
\right].
\end{aligned}
\end{equation}

where $(u)_{+}=\max(0,u)$, which penalizes real logits below $1$ and generated logits above $-1$. Averaging the fake logits in this dense discriminator yields the generator objective in Eq.~\ref{eq:feature_adv}. The discriminator is unconditional and must distinguish real and generated samples solely from visual evidence, while text conditioning remains enforced by the conditional generator and DMD objective.

In practice, sparse decoding reduces the activation memory for the frozen VAE and VFM. Generated rollouts are decoded without gradients during discriminator updates, while generator updates retain gradients only through latent neighborhoods. Activation checkpointing and stochastic spatial augmentation reduce memory and discourage reliance on fixed local statistics. Since the VAE decoder, VFM, and discriminator heads are training-only, the supervision adds no inference cost.



\subsection{Feature Space Exploration across Pretrained Backbones}
\label{sec:representation_backbones}

Our default instantiation uses a frozen DINOv2-S/14 encoder as its adversarial feature backbone~\citep{dinov2}. To study the effect of different adversarial feature space, we construct several comparison branches while keeping the causal student, on-policy DMD objective, autoregressive rollout, and inference architecture fixed. Their feature inputs can be summarized as
\begin{equation}
\mathcal{R}_{\omega,\ell}
=
\begin{cases}
\Phi_{\omega,\ell}
\left(
\mathcal{P}_{\omega}(\hat{\mathbf{x}}_m)
\right),
& \text{image encoder},\\[2mm]
\mathcal{V}_{\omega,\ell}
\left(
\mathcal{C}_{T,s}(\hat{\mathbf{x}})
\right),
& \text{video encoder},\\[2mm]
\mathcal{H}_{\omega,\ell}
\left(
q_{\tau}(\hat{\mathbf{z}};\boldsymbol{\epsilon}),
\tau,\varnothing
\right),
& \text{DiT backbone},
\end{cases}
\end{equation}
where $\omega$ identifies the frozen backbone, $\ell$ denotes an extracted feature level, $\mathcal{C}_{T,s}$ constructs a decoded clip of length $T$ and temporal stride $s$, and $q_{\tau}$ denotes the forward corruption process at timestep $\tau$. This common interface changes only the discriminator feature space while preserving the generator and distillation procedure as described in Sec.~\ref{sec:feature_discriminator}.

The image-level comparisons use VFM encoders DINOv2, DINOv3, and SigLIP2~\citep{dinov2,dinov3,siglip2}, with multi-level extraction and spatial discriminator heads. For video-native features, V-JEPA~2.1 and VideoMAE process decoded clips and project their tokens to discriminator width $C_h$ for temporal discrimination~\citep{vjepa21,videomae}. For diffusion features, real and generated VAE latents are corrupted with a shared timestep $\tau$ and noise realization $\boldsymbol{\epsilon}$ before entering the frozen DiT teacher shared with the real-score backbone. Features from layer set $\mathcal{S}_{\mathrm{DiT}}$ are projected to width $C_h$ and evaluated by local token-wise and pooled global heads. Sharing $\tau$ and $\boldsymbol{\epsilon}$ prevents the discriminator from exploiting corruption differences and instead focuses it on real--generated discrepancies. Unless otherwise specified, \textbf{Radian} denotes the DINOv2-S/14 configuration, with other encoders used in controlled ablations.

\section{Experiments}
\label{sec:experiments}

\subsection{Experimental Setup}
\label{sec:experimental_setup}

We evaluate our method across three autoregressive T2V settings: four-step (4-NFE) chunk-wise short-video generation, its long-horizon extension to one-minute videos, and one-step (1-NFE) frame-wise short-video generation. 

\noindent\textbf{Baselines.} For four-step chunk-wise generation, we compare our method with Self Forcing, Causal Forcing, and Salt~\citep{self_forcing,causal_forcing,salt}. We additionally include a locally implemented DiT-GAN that performs adversarial distillation over internal diffusion features, enabling a controlled comparison between different-based and VFM-based adversarial feature spaces. For one-step frame-wise generation, we compare against Causal Forcing++ and One-Forcing~\citep{causal_forcing_pp,one_forcing}, where One-Forcing is the baseline for DiT-GAN under 1-NFE settings. For long-video generation, we compare against Rolling Forcing \citep{rolling_forcing} and extend the generation results to one minute. For the baselines mentioned above, we re-evaluate the released checkpoints under our local environment.

\noindent\textbf{Benchmarks.} We evaluate four-step chunk-wise and one-step frame-wise short-video generation on VBench \citep{huang2024vbench}, reporting the normalized Total, Quality, and Semantic scores together with Dynamic Degree, Aesthetic Quality, and Imaging Quality. We additionally report VideoAlign \citep{videoalign} scores for visual quality (VQ), motion quality (MQ), text alignment (TA), and their aggregate score (Total). We evaluate the 60-second long-horizon video generation using VBench-Long \citep{huang2026vbenchpp}. For each generation setting, we evaluate all methods using matched prompts, resolutions, sampling settings, and scoring protocols, detailed in Appendix~\ref{app:implementation_details}.

\begin{table*}[t]
\centering
\caption{
Quantitative results under four-step chunk-wise (a), one-step frame-wise (b), and long-horizon autoregressive generation settings (c).
\#Params counts only the inference-time generator.
Darker green indicates a higher rank in each column.
}
\label{tab:main_quantitative}

\begin{subtable}[t]{\textwidth}
\centering
\caption{Four-step chunk-wise autoregressive generation results.}
\label{tab:main_short_4step}
\setlength{\tabcolsep}{3.4pt}
\resizebox{\linewidth}{!}{
\begin{tabular}{lcc*{13}{c}}
\toprule
& & &
\multicolumn{9}{c}{VBench $\uparrow$} &
\multicolumn{4}{c}{VideoAlign $\uparrow$} \\
\cmidrule(lr){4-12}\cmidrule(lr){13-16}
Method & \#Params & NFE
& \textbf{Total}
& Quality
& Semantic
& Subj. Cons.
& Dynamic
& Aesthetic
& Imaging
& Object
& Color
& VQ
& MQ
& TA
& \textbf{Total} \\
\midrule

Self Forcing \citep{self_forcing}
& 1.3B & 4
& \gcell{10}{0.8393}
& \gcell{10}{0.8471}
& \gcell{70}{0.8082}
& \gcell{15}{0.9378}
& \gcell{15}{0.6972}
& \gcell{10}{0.6745}
& \gcell{25}{0.6995}
& \gcell{10}{0.9440}
& \gcell{65}{0.8680}
& \gcell{70}{0.0695}
& \gcell{65}{0.0616}
& \gcell{15}{0.3683}
& \gcell{70}{0.4995} \\

Causal Forcing \citep{causal_forcing}
& 1.3B & 4
& \gcell{15}{0.8400}
& \gcell{25}{0.8495}
& \gcell{25}{0.8019}
& \gcell{10}{0.9310}
& \gcell{100}{0.8694}
& \gcell{15}{0.6766}
& \gcell{15}{0.6993}
& \gcell{65}{0.9544}
& \gcell{25}{0.8311}
& \gcell{15}{0.0072}
& \gcell{10}{-0.0951}
& \gcell{10}{0.3542}
& \gcell{10}{0.2663} \\

Salt \citep{salt}
& 1.3B & 4
& \gcell{65}{0.8434}
& \gcell{70}{0.8548}
& \gcell{10}{0.7982}
& \gcell{65}{0.9383}
& \gcell{25}{0.7926}
& \gcell{65}{0.6797}
& \gcell{65}{0.7034}
& \gcell{15}{0.9494}
& \gcell{10}{0.8291}
& \gcell{25}{0.0472}
& \gcell{15}{-0.0310}
& \gcell{70}{0.3774}
& \gcell{15}{0.3936} \\

DiT-GAN
& 1.3B & 4
& \gcell{25}{0.8411}
& \gcell{15}{0.8483}
& \gcell{100}{0.8122}
& \gcell{70}{0.9492}
& \gcell{10}{0.5954}
& \gcell{100}{0.6832}
& \gcell{10}{0.6984}
& \gcell{70}{0.9563}
& \gcell{70}{0.8713}
& \gcell{10}{-0.0242}
& \gcell{70}{0.0855}
& \gcell{25}{0.3715}
& \gcell{65}{0.4327} \\

\textbf{Radian} (CD init)
& 1.3B & 4
& \gcell{70}{0.8439}
& \gcell{100}{0.8550}
& \gcell{15}{0.7993}
& \gcell{25}{0.9380}
& \gcell{65}{0.7972}
& \gcell{25}{0.6792}
& \gcell{70}{0.7037}
& \gcell{25}{0.9508}
& \gcell{15}{0.8300}
& \gcell{65}{0.0521}
& \gcell{25}{-0.0296}
& \gcell{65}{0.3754}
& \gcell{25}{0.3979} \\

\textbf{Radian}
& 1.3B & 4
& \gcell{100}{0.8444}
& \gcell{65}{0.8541}
& \gcell{65}{0.8054}
& \gcell{100}{0.9504}
& \gcell{70}{0.8056}
& \gcell{70}{0.6801}
& \gcell{100}{0.7148}
& \gcell{100}{0.9570}
& \gcell{100}{0.8722}
& \gcell{100}{0.2134}
& \gcell{100}{0.2094}
& \gcell{100}{0.3805}
& \gcell{100}{0.8033} \\

\bottomrule
\end{tabular}}
\end{subtable}

\vspace{1.2mm}

\begin{subtable}[t]{\textwidth}
\centering
\caption{
One-step frame-wise autoregressive generation results.
$\dagger$ denotes First-Frame Enhancement, where the first latent frame uses four denoising steps and subsequent frames use one.
}
\label{tab:main_short_1step}
\setlength{\tabcolsep}{3.6pt}
\resizebox{\linewidth}{!}{
\begin{tabular}{lcc*{11}{c}}
\toprule
& & &
\multicolumn{7}{c}{VBench $\uparrow$} &
\multicolumn{2}{c}{VideoAlign $\uparrow$} \\
\cmidrule(lr){4-10}\cmidrule(lr){11-12}
Method & \#Params & NFE
& Total
& Quality
& Semantic
& Object
& App. Style
& Temp. Style
& Overall Cons.
& TA
& Total \\
\midrule

Causal Forcing++ \citep{causal_forcing_pp}
& 1.3B & $1^{\dagger}$
& \gcell{20}{0.8383}
& \gcell{20}{0.8473}
& \gcell{45}{0.8025}
& \gcell{20}{0.9473}
& \gcell{20}{0.7248}
& \gcell{45}{0.6908}
& \gcell{45}{0.7255}
& \gcell{45}{0.3115}
& \gcell{45}{-0.1813} \\

One Forcing \citep{one_forcing}
& 1.3B & $1^{\dagger}$
& \gcell{45}{0.8415}
& \gcell{90}{0.8533}
& \gcell{20}{0.7943}
& \gcell{45}{0.9568}
& \gcell{45}{0.7433}
& \gcell{20}{0.6907}
& \gcell{20}{0.7178}
& \gcell{20}{0.3168}
& \gcell{20}{0.0448} \\

\textbf{Radian}
& 1.3B & $1^{\dagger}$
& \gcell{90}{0.8417}
& \gcell{45}{0.8511}
& \gcell{90}{0.8040}
& \gcell{90}{0.9663}
& \gcell{90}{0.7546}
& \gcell{90}{0.7034}
& \gcell{90}{0.7267}
& \gcell{90}{0.4105}
& \gcell{90}{0.2563} \\

\bottomrule
\end{tabular}}
\end{subtable}

\vspace{1.2mm}

\begin{subtable}[t]{\textwidth}
\centering
\caption{
Long-horizon generation results on one-minute videos.
}
\label{tab:vbench_long}
\setlength{\tabcolsep}{4pt}
\renewcommand{\arraystretch}{1.05}
\resizebox{\linewidth}{!}{%
\begin{tabular}{lcc*{8}{c}}
\toprule
& & & \multicolumn{8}{c}{\textbf{VBench-Long} $\uparrow$} \\
\cmidrule(lr){4-11}
Method & \#Params & NFE
& Total
& Quality
& Semantic
& Subject Consist.
& Temp. Flicker
& Motion Smooth.
& Dynamic Degree
& Imaging Quality \\
\midrule

Rolling Forcing \citep{rolling_forcing}
& 1.3B & 5
& \gcell{15}{0.7805}
& \gcell{15}{0.8191}
& \gcell{90}{\textbf{0.6260}}
& \gcell{15}{0.9753}
& \gcell{15}{0.9871}
& \gcell{15}{0.9842}
& \gcell{15}{0.4532}
& \gcell{15}{0.7075} \\

\textbf{Radian}
& 1.3B & 4
& \gcell{90}{\textbf{0.8041}}
& \gcell{90}{\textbf{0.8511}}
& \gcell{15}{0.6160}
& \gcell{90}{\textbf{0.9790}}
& \gcell{90}{\textbf{0.9882}}
& \gcell{90}{\textbf{0.9864}}
& \gcell{90}{\textbf{0.6741}}
& \gcell{90}{\textbf{0.7194}} \\

\bottomrule
\end{tabular}}
\end{subtable}
\label{tab:main_table}
\vspace{-1em}
\end{table*}

\subsection{Main Results}
\label{sec:main_results}

\begin{figure*}[!t]
    \centering
    \includegraphics[width=\linewidth]{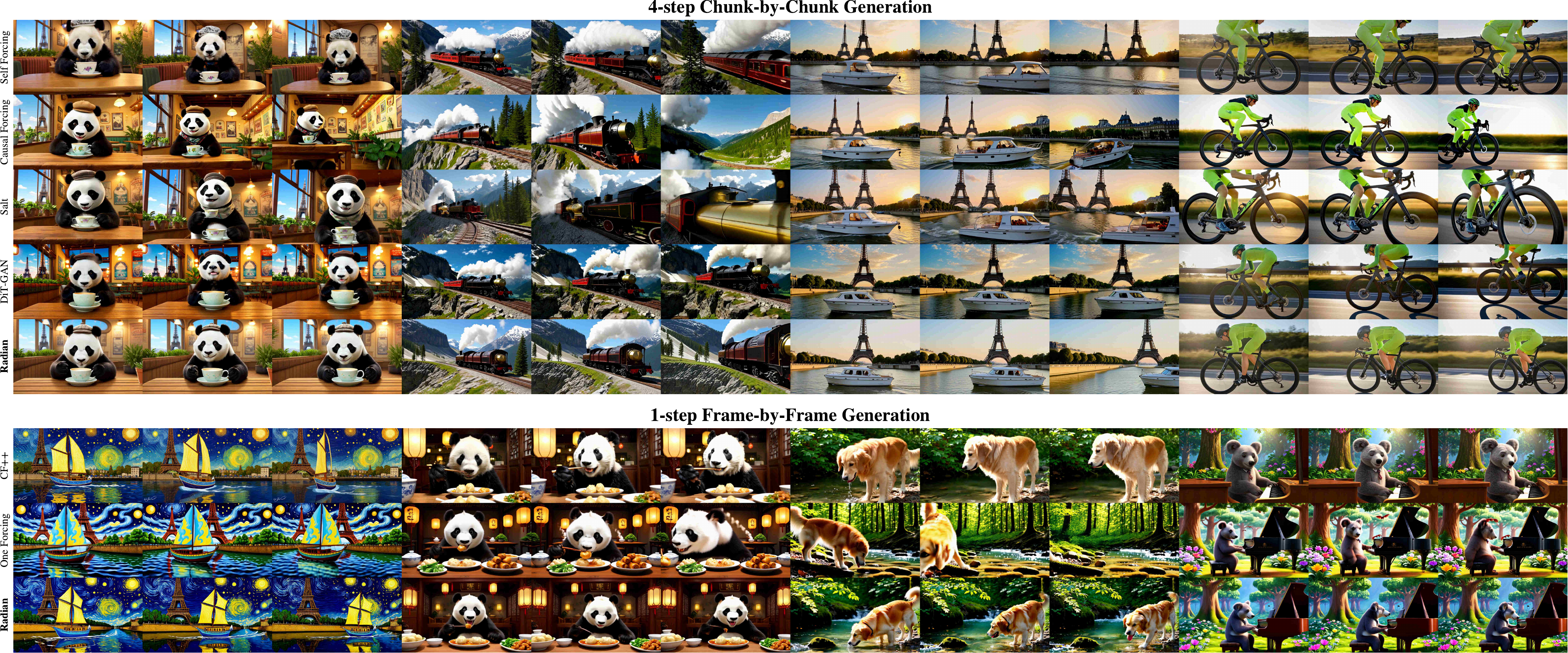}
    \caption{Selected short-video generation results using four-step chunk-wise generation and one-step frame-wise generation. More qualitative results and full prompts are provided in Appendix~\ref{sec:more_visualization_results}.}
    \label{fig:qualitative_41step}
\end{figure*}

\noindent\textbf{4-step Chunk-wise 5-Second Video Generation.} Tab.~\ref{tab:main_table}
(\subref{tab:main_short_4step}) reports VBench and VideoAlign results for four-step chunk-wise generation. We apply \textbf{Radian} to two ODE-initialized checkpoints: Self Forcing initialized from a bidirectional teacher and Causal Forcing from an autoregressive teacher (denoted as CD init). Under the same training recipe (8 B300 GPUs, 200 iterations for Stage~II, and 600 iterations for Stage~III), our method achieves the best performance with a \textit{VBench Total} of $0.8444$ and a \textit{VideoAlign Total} of $0.8033$. It consistently outperforms Self Forcing, Causal Forcing and Salt across the principal metrics under the same four-step sampling budget. As further illustrated in Fig.~\ref{fig:qualitative_41step} and Fig.~\ref{fig:more_chunk_wise}, it produces sharper details, more coherent structures, better prompt alignment, and higher-quality motion while suppressing undesirable modes observed in competing methods.

\noindent\textbf{1-step Frame-wise 5-Second Video Generation.} Tab.~\ref{tab:main_table}
(\subref{tab:main_short_1step}) reports VBench and VideoAlign results under the severely constrained one-step-per-frame setting ($1^{\dagger}$ NFE). \textbf{Radian} achieves a \textit{VBench Total} of $0.8417$, a \textit{Temporal Style} score of $0.7034$, and a \textit{VideoAlign Total} of $0.2563$, versus $0.0448$ for One Forcing, while also outperforming Causal Forcing++ in overall performance and temporal stability. Although One Forcing, a DiT-GAN baseline using internal DiT features, attains competitive aggregate VBench scores, its clips exhibit rapid flickering, unwanted zoom-ins, and structural deformations that inflate metrics despite degrading visual quality, as shown in Fig.~\ref{fig:qualitative_41step} and Fig.~\ref{fig:more_frame_wise}. In contrast, our method suppresses these artifacts and produces sharper frames, more stable structures, and coherent object and camera motion, indicating that decoded-output supervision in pretrained visual features provides a more reliable adversarial signal at $1^{\dagger}$ NFE.

\begin{figure*}[!t]
    \centering
    \includegraphics[width=\linewidth]{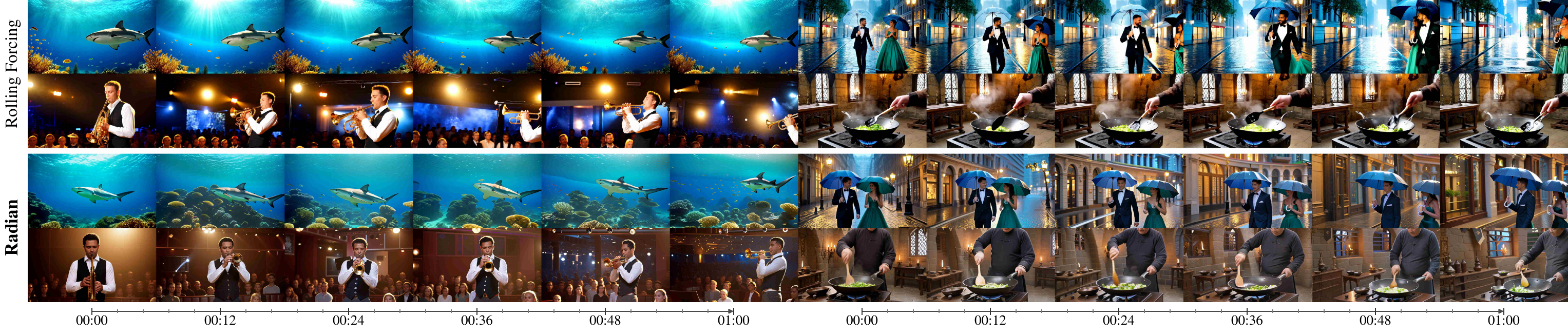}
    \caption{Selected long-video generation results using four-step chunk-wise generation. }
    \label{fig:qualitative_long}
    \vspace{-4mm}
\end{figure*}

\noindent\textbf{60-Second Long-Video Generation.} Tab.~\ref{tab:main_table}
(\subref{tab:vbench_long}) reports VBench-Long results for 957-frame videos of approximately one minute. \textbf{Radian} also establishes a new state of the art, achieving a \textit{VBench-Long Total} of $0.8041$, a \textit{VBench-Long Quality Score} of $0.8511$, a \textit{Dynamic Degree} of $0.6741$, and an \textit{Imaging Quality} score of $0.7194$. It significantly outperforms Rolling Forcing while using fewer denoising steps. As shown in Fig.~\ref{fig:qualitative_long} and Fig.~\ref{fig:more_long_video}, our method can better preserve visual details and coherent dynamics over long autoregressive rollouts of 60 seconds, whereas the baseline exhibits accumulated degradation or conservative motion. By adversarially suppressing low-quality feature modes at each autoregressive step, it reduces the local errors propagated through subsequent conditioning contexts, thereby limiting long-horizon error accumulation and improving temporal continuity. More qualitative results and full prompts are provided in Appendix~\ref{sec:more_visualization_results}.

\begin{table*}[!t]
\centering
\caption{
Ablation on different feature models for adversarial feature supervision.
}
\label{tab:ablation_vfm}
\setlength{\tabcolsep}{3pt}
\renewcommand{\arraystretch}{1.05}
\resizebox{\linewidth}{!}{%
\begin{tabular}{ll|*{12}{c}}
\toprule
Feature source & Feature model
& Total $\uparrow$
& Quality $\uparrow$
& Semantic $\uparrow$
& Multi-Obj. $\uparrow$
& Color $\uparrow$
& Consist. $\uparrow$
& Dynamic $\uparrow$
& Imaging $\uparrow$
& VQ $\uparrow$
& MQ $\uparrow$
& TA $\uparrow$
& VA Total $\uparrow$ \\
\midrule

Image VFM
& DINOv2
& \gcell{90}{0.8444}
& \gcell{90}{0.8541}
& \gcell{30}{0.8054}
& \gcell{10}{0.8343}
& \gcell{50}{0.8722}
& \gcell{20}{0.2662}
& \gcell{90}{0.8056}
& \gcell{90}{0.7148}
& \gcell{90}{0.2134}
& \gcell{70}{0.2094}
& \gcell{70}{0.3805}
& \gcell{90}{0.8033} \\

Image VFM
& DINOv3
& \gcell{40}{0.8401}
& \gcell{50}{0.8498}
& \gcell{10}{0.8012}
& \gcell{50}{0.8528}
& \gcell{70}{0.8729}
& \gcell{10}{0.2649}
& \gcell{70}{0.7685}
& \gcell{70}{0.7104}
& \gcell{70}{0.1462}
& \gcell{90}{0.2937}
& \gcell{30}{0.3607}
& \gcell{70}{0.8006} \\

Image VFM
& SigLIP2
& \gcell{30}{0.8391}
& \gcell{30}{0.8475}
& \gcell{20}{0.8052}
& \gcell{20}{0.8384}
& \gcell{30}{0.8662}
& \gcell{90}{0.2680}
& \gcell{40}{0.6343}
& \gcell{20}{0.6992}
& \gcell{50}{0.1156}
& \gcell{40}{0.1320}
& \gcell{50}{0.3751}
& \gcell{50}{0.6226} \\

Video VFM
& V-JEPA 2.1
& \gcell{10}{0.8360}
& \gcell{10}{0.8430}
& \gcell{50}{0.8078}
& \gcell{70}{0.8604}
& \gcell{20}{0.8627}
& \gcell{40}{0.2676}
& \gcell{30}{0.5991}
& \gcell{40}{0.7042}
& \gcell{20}{0.0498}
& \gcell{30}{0.1275}
& \gcell{90}{0.4119}
& \gcell{40}{0.5892} \\

Video VFM
& VideoMAE
& \gcell{20}{0.8372}
& \gcell{20}{0.8436}
& \gcell{70}{0.8115}
& \gcell{30}{0.8417}
& \gcell{90}{0.8787}
& \gcell{50}{0.2676}
& \gcell{10}{0.5213}
& \gcell{30}{0.7026}
& \gcell{40}{0.1000}
& \gcell{50}{0.1452}
& \gcell{20}{0.3436}
& \gcell{30}{0.5888} \\

Image + Video VFM
& DINOv2 + VideoMAE
& \gcell{70}{0.8434}
& \gcell{70}{0.8525}
& \gcell{40}{0.8070}
& \gcell{90}{0.8652}
& \gcell{10}{0.8545}
& \gcell{70}{0.2679}
& \gcell{50}{0.6593}
& \gcell{50}{0.7074}
& \gcell{30}{0.0831}
& \gcell{10}{0.0575}
& \gcell{10}{0.3079}
& \gcell{20}{0.4485} \\

Generative
& DiT-GAN
& \gcell{50}{0.8411}
& \gcell{40}{0.8483}
& \gcell{90}{0.8122}
& \gcell{40}{0.8432}
& \gcell{40}{0.8713}
& \gcell{30}{0.2674}
& \gcell{20}{0.5954}
& \gcell{10}{0.6984}
& \gcell{10}{-0.0242}
& \gcell{20}{0.0855}
& \gcell{40}{0.3715}
& \gcell{10}{0.4327} \\

\bottomrule
\end{tabular}%

}
\end{table*}



\begin{table*}[!t]
\centering

\begin{minipage}[t]{0.48\linewidth}
\centering
\captionof{table}{Ablation on DINOv2 encoder size.}
\label{tab:ablation_dinov2_size}
\vspace{-2mm}

\setlength{\tabcolsep}{2.6pt}
\renewcommand{\arraystretch}{0.80}
\setlength{\aboverulesep}{0.2ex}
\setlength{\belowrulesep}{0.2ex}
\scriptsize
\begin{tabular}{c|cccc}
\toprule
Size
& \shortstack{VBench Total} $\uparrow$
& Quality $\uparrow$
& Semantic $\uparrow$
& \shortstack{VA Total} $\uparrow$ \\
\midrule

S
& \gcell{90}{0.8444}
& \gcell{90}{0.8541}
& \gcell{90}{0.8054}
& \gcell{50}{0.8033} \\

B
& \gcell{10}{0.8322}
& \gcell{10}{0.8404}
& \gcell{10}{0.7993}
& \gcell{90}{0.8910} \\

L
& \pcell{0.8301}
& \pcell{0.8366}
& \pcell{0.8042}
& \pcell{0.3686} \\

G
& \gcell{50}{0.8387}
& \gcell{50}{0.8475}
& \gcell{50}{0.8035}
& \gcell{10}{0.3257} \\

\bottomrule
\end{tabular}
\end{minipage}
\hfill
\begin{minipage}[t]{0.50\linewidth}
\centering
\captionof{table}{Ablation on DINOv2 layer number.}
\label{tab:ablation_dino_layers}
\vspace{-2mm}

\setlength{\tabcolsep}{2.6pt}
\renewcommand{\arraystretch}{0.80}
\setlength{\aboverulesep}{0.2ex}
\setlength{\belowrulesep}{0.2ex}
\scriptsize
\begin{tabular}{c|cccc}
\toprule
\# Layers
& \shortstack{VBench Total} $\uparrow$
& Quality $\uparrow$
& Semantic $\uparrow$
& \shortstack{VA Total} $\uparrow$ \\
\midrule

1
& \gcell{10}{0.8182}
& \gcell{10}{0.8206}
& \gcell{90}{0.8085}
& \gcell{30}{0.6747} \\

2
& \gcell{30}{0.8283}
& \gcell{30}{0.8389}
& \gcell{10}{0.7859}
& \gcell{10}{0.6586} \\

4
& \gcell{90}{0.8444}
& \gcell{90}{0.8541}
& \gcell{50}{0.8054}
& \gcell{90}{0.8033} \\

8
& \gcell{50}{0.8381}
& \gcell{50}{0.8469}
& \gcell{30}{0.8026}
& \gcell{50}{0.7228} \\

\bottomrule
\end{tabular}
\end{minipage}

\vspace{-1em}
\end{table*}

\subsection{Adversarial Distillation in Different Representation Spaces}
\label{sec:representation_comparison}

\noindent\textbf{Image Representation Feature Spaces.}
As shown in Tab.~\ref{tab:ablation_vfm}, the three image VFMs induce adversarial gradients with different magnitudes and directions, leading to different optimization strengths under the same objective. DINOv2-S provides a moderate and stable signal, achieving the best overall balance with a VBench \textit{Total} of $0.8444$ and a VideoAlign \textit{Total} of $0.8033$. DINOv3-S produces larger gradients and achieves the highest MQ score of $0.2937$, although its aggressive updates are harder to stabilize. SigLIP2 instead provides a more semantics-oriented signal, achieving the highest \textit{Overall Consistency} score of $0.2680$ while being less sensitive to fine-grained rendering and motion artifacts. These differences arise from each encoder's feature scale, token geometry, and input Jacobian; thus, identical discriminator learning rates and adversarial weights do not imply equal supervision strength and require separate calibration.


\noindent\textbf{Temporally Structured Video Features.}
Video VFMs encode cross-frame dependencies and tend to regularize uncertain temporal changes, improving consistency partly by reducing motion amplitude. As shown in Tab.~\ref{tab:ablation_vfm}, V-JEPA 2.1 achieves the highest TA score of $0.4119$ while its Dynamic score drops to $0.5991$, reflecting its preference for predictable and temporally persistent content. VideoMAE preserves richer local semantic and appearance cues, reaching a Semantic score of $0.8115$ and a Color score of $0.8787$, but at the cost of an even lower Dynamic score of $0.5213$. Combining DINOv2 with VideoMAE provides a better trade-off between artifact-sensitive spatial supervision and complementary temporal regularization, raising the VBench \textit{Total} to $0.8434$, Dynamic to $0.6593$, and Multi-Object to $0.8652$. These results suggest that video representations are most effective as complementary temporal priors rather than replacements for DINOv2 supervision.


\noindent\textbf{Complementarity with DMD Gradients.}
DiT-GAN exhibits different optimization behavior from VFM-based adversarial supervision. As shown in Tab.~\ref{tab:ablation_vfm}, it achieves the strongest Semantic score of $0.8122$, plausibly because diffusion-teacher features align with latent denoising and retain task-relevant generative semantics. Nevertheless, its VideoAlign profile remains weak: VQ, MQ, and TA are $-0.0242$, $0.0855$, and $0.3715$, yielding the worst VideoAlign \textit{Total} of $0.4327$. Our gradient-direction analysis suggests that DiT-GAN gradients are strongly coupled with DMD, reinforcing directions already present in the distillation objective. In contrast, gradients from DINO-, SigLIP-, V-JEPA-, and VideoMAE-based discriminators are approximately orthogonal to DMD. This suggests that representation-space adversarial supervision introduces complementary perceptual, semantic, or temporal directions rather than merely strengthening distillation.
More qualitative comparisons are provided in Appendix~\ref{app:additional_qualitative}.


\subsection{Ablation Study}
\label{sec:ablation}

\noindent\textbf{Ablation on Encoder Size.}
We examine whether scaling the frozen DINOv2 encoder improves adversarial distillation. As shown in Tab.~\ref{tab:ablation_dinov2_size}, larger encoders do not yield monotonic gains, while DINOv2-S/14 provides the best overall balance in VBench and VideoAlign. DINOv2-G/14 reaches $0.8387$ and $0.3257$, respectively, whereas DINOv2-B/14 achieves a higher VideoAlign \textit{Total} of $0.8910$ but a lower VBench \textit{Total} of $0.8322$. This is consistent with ADD, where DINOv2 ViT-S also outperforms ViT-L as the discriminator backbone~\citep{ADD}. Larger ViTs may discard local appearance cues through increasingly invariant semantic features, while higher feature dimensionality and discriminator capacity can yield noisier gradients. Thus, a stronger encoder does not necessarily provide a better adversarial feature space.

\noindent\textbf{Ablation on Feature Layer Selection.}
We further ablate the number and positions of DINOv2 layers used by the discriminator. In Tab.~\ref{tab:ablation_dino_layers}, the one-, two-, four-, and eight-layer settings use blocks $\{11\}$, $\{6,11\}$, $\{6,8,10,11\}$, and $\{2,4,6,7,8,9,10,11\}$, respectively. Specifically, VBench \textit{Total} increases from $0.8182$ to $0.8283$, peaks at $0.8444$ with four layers, and drops to $0.8381$ with eight, indicating diminishing returns. Too few layers provide insufficient multi-level supervision, while too many introduce redundant low-level cues. In particular, the final block overemphasizes invariant semantics and weakens local details, whereas early layers can overemphasize textures. The four layers therefore balance structural, semantic, and local appearance evidence without a high complexity.
Visual results for the ablations are provided in Appendix~\ref{app:additional_qualitative}.

\section{Conclusion}
\label{conclusion}

This paper introduced \textbf{Radian}, a representation adversarial distillation framework for few-step autoregressive video generation. While DMD transfers the diffusion teacher's generative prior in latent space, Radian complements it with real-data adversarial supervision over multi-level features from a frozen visual foundation model, directly improving decoded video quality without additional inference cost. Experiments across short, one-step, or long-horizon generation demonstrate improvements in visual fidelity, motion quality, and autoregressive stability. Our analysis further shows that different representation spaces induce distinct optimization signals: image VFMs differ in gradient strength and emphasize different perceptual or semantic cues, while video-native representations provide temporal regularization, often with reduced motion diversity. Moreover, external VFM gradients are largely complementary to DMD, whereas diffusion-internal adversarial features are more strongly coupled with the original distillation objective. These findings highlight representation-space supervision as an effective way to refine few-step autoregressive video generation quality.

\bibliography{iclr2027_conference}
\bibliographystyle{iclr2027_conference}

\clearpage
\appendix

\begin{center}
  {\Large\bfseries Appendix\par}
\end{center}
\phantomsection
\label{app:contents}
\vspace{0.65em}

\begingroup
\color{black}
\hypersetup{hidelinks}
\setlength{\fboxsep}{8pt}
\setlength{\fboxrule}{0.4pt}
\noindent\fcolorbox{black}{white}{%
\begin{minipage}{\dimexpr\linewidth-2\fboxsep-2\fboxrule\relax}
\raggedright
{\normalsize Contents\par}
\vspace{0.45em}
\small
\setlength{\parskip}{0.12em}

\newcommand{\radianappsectionentry}[2]{%
  \noindent{\hyperref[#1]{%
    \makebox[2.2em][l]{\textbf{\ref*{#1}}}\textbf{#2}}}%
  \hfill\hyperref[#1]{\textbf{\pageref*{#1}}}\par
}

\newcommand{\radianappsubsectionentry}[2]{%
  {\footnotesize\noindent\hspace*{1.2em}%
  {\hyperref[#1]{%
    \makebox[3.2em][l]{\ref*{#1}}#2}}%
  \nobreak\leaders\hbox to 0.55em{\hss.\hss}\hfill\nobreak
  \hyperref[#1]{\pageref*{#1}}\par}%
}

{\footnotesize\itshape Main text\par}
\vspace{0.15em}

\radianappsectionentry{sec:introduction}{Introduction}
\radianappsectionentry{sec:related_work}{Related Work}
\radianappsectionentry{sec:method}{Method}
\radianappsubsectionentry{sec:preliminaries}{Preliminaries}
\radianappsubsectionentry{sec:on_policy_rad}{Representation Adversarial Distillation for Causal Video Generation}
\radianappsubsectionentry{sec:feature_discriminator}{Multi-Level Feature Discrimination over Decoded Outputs}
\radianappsubsectionentry{sec:representation_backbones}{Feature Space Exploration across Pretrained Backbones}
\radianappsectionentry{sec:experiments}{Experiments}
\radianappsubsectionentry{sec:experimental_setup}{Experimental Setup}
\radianappsubsectionentry{sec:main_results}{Main Results}
\radianappsubsectionentry{sec:representation_comparison}{Adversarial Distillation in Different Representation Spaces}
\radianappsubsectionentry{sec:ablation}{Ablation Study}
\radianappsectionentry{conclusion}{Conclusion}

\vspace{0.5em}
{\footnotesize\itshape Appendix\par}
\vspace{0.15em}

\radianappsectionentry{app:implementation_details}{Detailed Implementation and Evaluation}
\radianappsubsectionentry{app:rad_notation}{Notation}
\radianappsubsectionentry{app:rad_evaluation}{Evaluation Protocol}
\radianappsubsectionentry{app:rad_metrics}{Metric Definitions}
\radianappsubsectionentry{app:rad_sampling}{Models and Sampling}
\radianappsubsectionentry{app:rad_training}{Training Configuration}
\radianappsubsectionentry{app:rad_vfm}{Representation Configurations and Ablations}
\radianappsectionentry{sec:more_visualization_results}{Additional Visualizations and Prompt Details}
\radianappsubsectionentry{app:additional_qualitative}{Additional Qualitative Comparisons}
\radianappsubsectionentry{app:main_visual_prompts}{Prompts for Main-Text Visualizations}

\end{minipage}%
}
\endgroup
\vspace{0.75em}

\section{Detailed Implementation and Evaluation}
\label{app:implementation_details}

This appendix describes notation, evaluation, sampling, training, and the image- and video-representation configurations used in Radian.

\subsection{Notation}
\label{app:rad_notation}

Table~\ref{tab:rad_notation} collects the principal symbols. Diffusion time $t$ is distinct from optimization iteration $u$; $z$ denotes generator noise, whereas $\hat z$ denotes its generated clean latent sequence. Transformer block indices are zero-based.

\begin{table}[htb]
\centering
\small
\caption{Principal notation. }
\label{tab:rad_notation}
\setlength{\tabcolsep}{5pt}
\renewcommand{\arraystretch}{1.08}
\begin{tabular}{@{}p{0.27\linewidth}p{0.69\linewidth}@{}}
\toprule
Symbol & Meaning\\
\midrule
$c$, $z$ & Text condition and initial Gaussian generator input.\\
$G_\theta$, $p_\theta^{\mathrm{AR}}$ & Causal student and its autoregressive output distribution.\\
$\hat z^{(b)}$, $\hat z^{(<b)}$ & Generated latent block and its preceding context.\\
$x_0$, $x_t$ & Clean and diffusion-corrupted video latents.\\
$s_r$, $s_\phi$ & Frozen teacher score and trainable fake-score model.\\
$\mathcal D_{\mathrm{VAE}}$ & Frozen latent-to-RGB decoder.\\
$x$, $\hat x$ & Real and generated RGB inputs to representation supervision.\\
$\Phi$, $\mathcal P$, $\mathcal R_\Phi$ & Frozen visual encoder, input preprocessing, and selected multi-level representation.\\
$\mathcal S$, $F_\ell$ & Selected transformer blocks and level-$\ell$ feature tokens.\\
$H_\psi$, $h_{\psi,\ell}$ & Trainable representation discriminator and its level-specific head.\\
$T_z$, $T_x$, $M$ & Latent length, decoded RGB length, and sampled image-frame count.\\
$\mathcal L_{\mathrm{DMD}}$, $\mathcal L_{\mathrm{adv}}^\Phi$, $\mathcal L_D$ & Distribution-matching, generator adversarial, and discriminator losses.\\
$\lambda_{\mathrm{DMD}}$, $\lambda_{\mathrm{adv}}$ & Generator objective weights.\\
$\operatorname{sg}(\cdot)$, $J_f$ & Stop-gradient operation and Jacobian of mapping $f$.\\
\bottomrule
\end{tabular}
\end{table}

\subsection{Evaluation Protocol}
\label{app:rad_evaluation}

\textbf{Shared prompts and seeds.}
All quantitative experiments use the same randomly-chosen 15 inference seeds as the four-step five-second evaluation: 2, 4, 1805718, 3367341, 3718700, 5036436, 5279402, 5450779, 5748298, 7252721, 9274039, 9539461, 5918269, 7730833, and 2436109. This shared set applies to every compared method in the four-step chunk-wise, one-step frame-wise, and one-minute experiments, as well as all feature-model, encoder-size (including DINOv2-L), and feature-depth ablations. These are inference seeds for fixed model weights, rather than independent training runs. The set was randomly selected from a larger seed pool for further benchmarking. We additionally conducted training runs with different random seeds and observed consistent trends, indicating that the reported improvements are not specific to a particular training seed. In separate internal experiments, we also further applied Radian to a proprietary video generation model and observed consistent improvements, suggesting that the proposed approach extends beyond the Wan2.1-1.3B setting.

Short-video evaluation uses 946 prompts at $832\times480$ resolution and 16 FPS. Generation uses the shared extended prompts, while scoring metadata retains their original benchmark descriptions. Within each evaluation setting, methods use the same prompts and seed set. We compute each metric with its benchmark evaluator for each seed and report the arithmetic mean over the 15 seeds. Qualitative examples are illustrative selections, not additional benchmark averages.

\textbf{One-minute evaluation.}
Long-video comparisons use the same 946 prompts and 15 inference seeds, generating 957 RGB frames at $832\times480$ and 16 FPS. VBench-Long evaluates the complete generated videos.

\subsection{Metric Definitions}
\label{app:rad_metrics}

\textbf{VBench.}
VBench~\citep{huang2024vbench} comprises seven quality and nine semantic dimensions. We apply its fixed reference bounds and dimension weights to obtain Quality $Q$ and Semantic $S$, and compute
\begin{equation}
 \mathrm{VBench\ Total}=\frac{4Q+S}{5}.
 \label{eq:rad_vbench_aggregate}
\end{equation}
All 16 dimensions contribute to these aggregates, regardless of which columns appear in a displayed table. The quality group comprises subject consistency, background consistency, temporal flickering, motion smoothness, dynamic degree, aesthetic quality, and imaging quality. The semantic group comprises object class, multiple objects, human action, color, spatial relationship, scene, appearance style, temporal style, and overall consistency. Within each group, we use the benchmark's weighted average of normalized dimension scores, not an unweighted average of the displayed columns. Dimension normalization uses $\widetilde r_d=\operatorname{clip}_{[0,1]}((r_d-a_d)/(b_d-a_d))$, where $a_d,b_d$ are the benchmark reference bounds. The feature-model ablation reports native style and overall-consistency scores, whereas the one-step main table uses their benchmark-normalized counterparts; these scales should not be compared directly.

\textbf{VideoAlign.}
VideoAlign~\citep{videoalign} measures visual quality (VQ), motion quality (MQ), and text alignment (TA). Our evaluator applies the reward checkpoint's fixed standardization:
\begin{equation}
 A_k=\frac{r_k-\mu_k}{\sigma_k},\qquad
 \mathrm{VA\ Total}=A_{\mathrm{VQ}}+A_{\mathrm{MQ}}+A_{\mathrm{TA}}.
 \label{eq:rad_videoalign}
\end{equation}
In VQ/MQ/TA order, the means are $(3.6757,1.1646,2.8105)$ and standard deviations are $(2.2476,1.3811,2.5121)$. These constants do not depend on the methods being compared. Scores are not probabilities or bounded by $[0,1]$: negative components and totals above one are valid. The evaluator samples at a nominal 2 FPS with at most 200,704 pixels per sampled frame.

\textbf{Long-video metrics.}
VBench-Long~\citep{huang2026vbenchpp} uses its long-video preprocessing and dimension-specific routines, including the configured static-content filter for temporal flickering. It does not score only the first five seconds. Dynamic Degree quantifies detected movement, not necessarily plausible motion; consistency can also increase when movement decreases. We therefore interpret these measures alongside imaging quality and qualitative videos, and report short- and long-video aggregates separately.

\subsection{Models and Sampling}
\label{app:rad_sampling}

\textbf{Networks and latent geometry.}
The student is a causal Wan2.1-T2V-1.3B model~\citep{wan}. A frozen bidirectional Wan2.1-T2V-14B supplies teacher predictions, and a trainable 1.3B fake-score model estimates the student distribution. The text encoder, VAE, and visual backbone are frozen. Only the student, fake-score network, and discriminator heads/projections are optimized. At inference, the teacher, fake-score network, visual backbone, and discriminator are removed.

The causal VAE maps $T_z$ latent frames to $T_x=1+4(T_z-1)$ RGB frames: 21 latents yield 81 frames, and 240 yield 957. The latent grid has 16 channels and spatial size $104\times60$. The decoder uses a temporal compression factor of four with causal context and a special first-frame boundary.

\begin{table}[t]
\centering
\small
\caption{Sampling settings at $832\times480$, 16 FPS. Schedules are indices before scheduler warping.}
\label{tab:rad_sampling}
\setlength{\tabcolsep}{4pt}
\resizebox{\linewidth}{!}{%
\begin{tabular}{lccc}
\toprule
Setting & Four-step short & One-step short & One-minute\\
\midrule
Latent / RGB frames & 21 / 81 & 21 / 81 & 240 / 957\\
Latent frames per block & 3 & 1 & 3\\
Denoising schedule & $\{1000,750,500,250\}$ & $\{1000\}$ & $\{1000,750,500,250\}$\\
First-block exception & None & Four-step enhancement & None\\
Student inference CFG & Disabled & Disabled & Disabled\\
Evaluation seeds & 15 & 15 & 15\\
\bottomrule
\end{tabular}}
\end{table}

\textbf{Denoising and guidance.}
Four-step sampling uses scheduler shift 5.0. Teacher guidance is applied during distillation, while student inference uses only the conditional branch. The implementation's guidance setting 3.0 means $\widehat x_r^{\mathrm{cfg}}=\widehat x_r(c)+ 3(\widehat x_r(c)-\widehat x_r(\varnothing))$; its conditional coefficient is therefore four. Fake-score predictions are conditional and use no additional guidance term.

The frame-wise model starts from the causal ODE initialization used by Causal Forcing++~\citep{causal_forcing_pp}. First-Frame Enhancement uses four steps for the first latent and one for each subsequent latent. Accordingly, $1^\dagger$ denotes the steady-state per-frame denoising budget; a 21-latent clip uses 24 denoising evaluations in total before cache-related operations.

\textbf{Rolling generation.}
The long-video experiment places the four-step Radian model into the rolling inference implementation of Rolling Forcing~\citep{rolling_forcing}, without additional long-video training. A staggered window of four three-latent blocks advances through the sequence while detached key/value caches retain clean context. The full 240-latent sequence is decoded into RGB. We report NFE based on denoising evaluations, with cache-writing passes and VAE decoding treated separately.

\subsection{Training Configuration}
\label{app:rad_training}

\textbf{Training data.}
All experiments use the same 40K captioned-video training collection. Real clips are uniformly sampled to 81 frames, resized to $832\times480$, and represented in $[-1,1]$ for VAE processing. Representation inputs are mapped to $[0,1]$ before encoder-specific normalization. Real and generated videos are sampled independently, with image sampling aligned by normalized temporal positions. The DMD and GAN objectives use separate on-policy rollouts.

\textbf{DMD-only continuation as reference.}
Self Forcing~\citep{self_forcing} and Causal Forcing~\cite{causal_forcing} serve as the DMD-based reference models for the two initialization routes. They do not use representation-space adversarial supervision, whereas Radian augments the corresponding causal generator with discriminator calibration followed by joint DMD and representation-adversarial refinement. We therefore use these baselines to assess the effect of introducing representation-space adversarial supervision.

\textbf{Stage I: causal ODE initialization.}
Stage I uses a causal ODE initialization of the pretrained foundation model. The main chunk-wise route uses the Self Forcing initialization~\citep{self_forcing}; the CD-initialized variant loads \texttt{causal\_cd.pt}. These provide alternative starting weights for subsequent Radian post-training.

The inherited ODE initialization regresses clean trajectory targets from intermediate noisy states under causal context:
\begin{equation}
 \mathcal L_{\mathrm{ODE}}=
 \mathbb E\left\|
 G_\theta(z_\tau^{\mathrm{ODE}},\tau,c;z_0^{\mathrm{ODE},<b})
 -z_0^{\mathrm{ODE},b}\right\|_2^2.
 \label{eq:rad_ode_init}
\end{equation}
Here $\tau$ is a selected trajectory time and $b$ is the predicted block. This objective is used during initialization. The subsequent on-policy stage conditions on the student's generated context.

\textbf{Stage II: on-policy adaptation and discriminator calibration.}
Stage II runs for 200 outer iterations. DMD and discriminator training are active throughout. Generator-side adversarial supervision remains disabled throughout Stage II, while the discriminator is calibrated using real videos and detached student rollouts. This stage adapts the initialized student to its generated context while calibrating the discriminator for subsequent joint refinement.

\textbf{Stage III: low-learning-rate joint refinement.}
Stage III loads the resulting generator weights and continues training with lower learning rates, using
\begin{equation}
 \mathcal L_G(u)=\lambda_{\mathrm{DMD}}\mathcal L_{\mathrm{DMD}}
       +g_u\lambda_{\mathrm{adv}}\mathcal L_{\mathrm{adv}}^\Phi,
 \qquad \lambda_{\mathrm{DMD}}=\lambda_{\mathrm{adv}}=1.
 \label{eq:rad_joint_appendix}
\end{equation}
Here $g_u$ denotes the scheduled adversarial activation and equals one throughout Stage III. The default continuation uses the local-step-700 EMA model. Stage III initializes from the Stage-II generator and calibrated discriminator, and jointly optimizes DMD and representation-adversarial objectives. Table~\ref{tab:rad_optim_common} summarizes the default optimization settings.

\begin{table}[t]
\centering
\small
\caption{Default chunk-wise optimization. Learning rates refer to trainable networks; the visual encoder remains frozen.}
\label{tab:rad_optim_common}
\setlength{\tabcolsep}{4pt}
\renewcommand{\arraystretch}{1.02}
\resizebox{0.65\linewidth}{!}{%
\begin{tabular}{@{}lccc@{}}
\toprule
Setting & Generator & Fake-score & Discriminator\\
\midrule
Optimizer & AdamW & AdamW & AdamW\\
$(\beta_1,\beta_2)$ & $(0,0.999)$ & $(0,0.999)$ & $(0,0.95)$\\
Weight decay & 0.01 & 0.01 & 0\\
Stage II learning rate & $2\times10^{-6}$ & $4\times10^{-7}$ & $3\times10^{-5}$\\
Stage III learning rate & $5\times10^{-7}$ & $4\times10^{-7}$ & $3\times10^{-6}$\\
Gradient-norm clipping & 10 & 10 & 10\\
EMA decay & 0.99 & Not used & Not used\\
\bottomrule
\end{tabular}}
\end{table}

\textbf{On-policy DMD.}
The student generates autoregressive rollouts conditioned on its own history. The frozen teacher and trainable fake-score model process the same diffusion-corrupted student latents, and their prediction difference provides the DMD update direction. The update direction is detached when optimizing the student, while the fake-score model is trained separately with a flow-denoising objective on detached student samples. The DMD and representation-adversarial objectives use separate on-policy rollouts sampled from the same student.

\textbf{Batching and hardware.}
Training follows a five-iteration update cycle: the generator is updated once every five outer iterations, while fake-score and discriminator updates occupy the intervening iterations. The default chunk-wise run uses eight workers, one video per worker, and eight DMD accumulation microbatches, resulting in 64 videos per DMD update. The GAN branch uses one video per worker and eight sampled frames per video, resulting in eight videos per update. Losses are mean-reduced within each branch.

All final experiments run on NVIDIA B300 GPUs. The default configuration uses eight GPUs, bfloat16 mixed precision, hybrid FSDP, and activation checkpointing. Checkpointed frozen-encoder and VAE forwards retain input gradients during generator updates.

\textbf{Frame-wise post-training.}
The one-step recipe uses 500 DMD-only iterations followed by 100 iterations of discriminator calibration before joint training. At joint activation, the generator learning rate changes from $2\times10^{-6}$ to $5\times10^{-7}$; the discriminator and fake-score learning rates are $3\times10^{-6}$ and $4\times10^{-7}$, respectively. One DMD microbatch per worker gives an effective batch size of eight videos. The DMD stream uses text prompts, while the adversarial stream uses captions associated with real videos.

\subsection{Representation Configurations and Ablations}
\label{app:rad_vfm}

\textbf{Image sampling and preprocessing.}
The image branch samples eight of the 21 latent positions uniformly without replacement. Position $i$ aligns to real RGB index $\operatorname{round}(i(T_r-1)/(T_z-1))$, or $4i$ for 81 RGB frames. Here $T_r$ is the real clip's RGB length. Selected one-latent decoding chunks provide a sparse approximation of full-sequence decoding during training. DINOv2-S/14~\citep{dinov2} uses area resizing to $518\times518$ and ImageNet normalization, producing a $37\times37$ patch grid. With probability 0.5, a square crop of side 64, 128, or 192 pixels is applied before resizing. Real and generated inputs share the same augmentation distribution.

ImageNet channel means are $(0.485,0.456,0.406)$ and standard deviations are $(0.229,0.224,0.225)$. The crop operates in RGB coordinates before encoder resizing. Real and generated videos are sampled independently, while their frames are aligned by normalized temporal position. The discriminator therefore captures distribution-level differences across real and generated samples. Cached generated samples are detached before being reused for discriminator training.

\textbf{Dense discriminator.}
The default discriminator uses transformer blocks $\{6,8,10,11\}$ together with the patch-input feature after positional embedding. At each level, the CLS readout is broadcast-added to the patch tokens. Each level has an independent head composed of spectrally normalized Conv1d layers, local normalization, LeakyReLU with slope 0.2, and a residual kernel-9 block with circular padding and scaling $1/\sqrt{2}$. A final kernel-1 layer predicts dense logits. Convolutions operate over flattened token sequences. Local normalization uses virtual groups of eight, and channel projections are trainable where required.

For real and generated logits $d^{\mathrm{real}}$ and $d^{\mathrm{fake}}$,
\begin{align}
 \mathcal L_D
 &=\mathbb E\langle(1-d^{\mathrm{real}})_+\rangle
   +\mathbb E\langle(1+d^{\mathrm{fake}})_+\rangle,\\
 \mathcal L_{\mathrm{adv}}^\Phi
 &=-\mathbb E\langle d^{\mathrm{fake}}\rangle .
 \label{eq:rad_hinge_appendix}
\end{align}
Here $(a)_+=\max(a,0)$ and brackets denote averaging over concatenated logits. Levels receive equal weight when their token counts match; otherwise, the reduction is token-weighted. The discriminator is unconditional.

\textbf{Gradient through a frozen representation.}
Write $x_\theta=\mathcal D_{\mathrm{VAE}}(G_\theta(z,c))$ and $\mathcal R_\Phi=\Phi_{\mathcal S}\circ\mathcal P$. For a fixed discriminator and sampled preprocessing transform, let $h_\psi$ denote the mean dense logit. The chain rule gives
\begin{equation}
 \nabla_\theta\mathcal L_{\mathrm{adv}}^\Phi
 =-\mathbb E\!\left[
 J_{G_\theta}^{\mathsf T}J_{\mathcal D_{\mathrm{VAE}}}^{\mathsf T}
 J_{\mathcal P}^{\mathsf T}J_{\Phi_{\mathcal S}}^{\mathsf T}
 \nabla_F h_\psi(F)\right],
 \quad F=\mathcal R_\Phi(x_\theta).
 \label{eq:rad_feature_gradient}
\end{equation}
Although the encoder parameters are frozen, gradients propagate through its input Jacobian. The adversarial update is therefore shaped by decoded appearance and the encoder's feature sensitivity, while the DMD update is determined by teacher and fake-score predictions. These two gradient pathways provide distinct sources of supervision to the generator. During discriminator updates, inputs are detached; during generator updates, discriminator heads are fixed while gradients remain active through the VAE, preprocessing, and frozen representation encoder.

When a head normalizes groups of inputs, $F$ in Eq.~\ref{eq:rad_feature_gradient} denotes the stacked group, and the corresponding Jacobian includes cross-input dependencies. For randomized crops, the gradient expression is evaluated for each sampled transform and averaged over the preprocessing distribution.

\textbf{Complementary Gradients.}
In an auxiliary VideoMAE-based run, measurements over \textbf{127} generator updates yield a GAN--DMD gradient cosine similarity of $\mathbf{-0.0234 \pm 0.0534}$ (mean $\pm$ standard deviation), indicating approximately orthogonal directions on average in this diagnostic. The median adversarial-to-DMD gradient-norm ratio is $\mathbf{0.8381}$, showing that the adversarial signal has a comparable magnitude. These loss-weighted, pre-clipping measurements support directional complementarity in the examined run, although they do not establish universal orthogonality across representation backbones or, by themselves, demonstrate improved optimization.

\begin{table*}[t]
\centering
\caption{Representation designs with frozen backbones. $P$ denotes the patch-input branch; channel arrows indicate trainable projections. Image inputs are written as frame count $\times$ spatial resolution. Video inputs use 32 frames, short side 256, and RGB stride two. The mixed variant combines DINOv2-S/14 with VideoMAE; its video head uses spatial pooling.}
\label{tab:rad_vfm_design}
\setlength{\tabcolsep}{4.5pt}
\renewcommand{\arraystretch}{1.16}
\footnotesize
\resizebox{\linewidth}{!}{%
\begin{tabular}{@{}llccl@{}}
\toprule
Backbone & Input & Feature levels & Channels & Readout / heads\\
\midrule
DINOv2-S/14 & $8\times518^2$ & $P+\{6,8,10,11\}$ & $384$ & CLS + dense\\
DINOv3-S/16 & $8\times512^2$ & $P+\{2,5,8,11\}$ & $384$ & Norm. patch mean + dense\\
SigLIP2-B/16 & $8\times512^2$ & $P+\{6,8,10,11\}$ & $768\rightarrow512$ & Norm. patch mean + dense\\
\addlinespace[2pt]
V-JEPA 2.1-B (distilled) & Video clip & $P+\{6,8,10,11\}$ & $768\rightarrow512$ & Spatiotemporal dense\\
VideoMAE & Video clip & $P+\{6,8,10,11\}$ & $768\rightarrow512$ & Spatiotemporal dense\\
\addlinespace[2pt]
DINOv2 + VideoMAE & Images + clip & Image levels + video output & $384;\ 768\rightarrow512$ & Image dense + temporal\\
DiT-GAN (Wan-14B) & Noisy latents & $\{21,28,35,39\}$ & $5120\rightarrow384$ & Frame-local + video-pooled\\
\bottomrule
\end{tabular}}
\end{table*}

\textbf{Image-encoder adaptations.}
DINOv3~\citep{dinov3} uses bilinear resizing and ImageNet normalization; SigLIP2~\citep{siglip2} uses bicubic resizing and channel mean/standard deviation 0.5. Both use normalized features and a final patch-mean readout in place of the default CLS readout. Selected intermediate features receive affine-free LayerNorm. SigLIP2's final output retains its pretrained post-normalization, and its 768-channel tokens are projected to 512 channels before discrimination. DINOv3 uses 384-channel heads without the default local normalization. The image feature heads preserve spatial tokens instead of reducing each frame to a single classification embedding. Table~\ref{tab:rad_vfm_design} summarizes the feature and head designs. Since input geometry and head adaptations vary across model families, these experiments evaluate the resulting representation-supervision configurations as complete design choices.

\textbf{Video and mixed representations.}
V-JEPA 2.1~\citep{vjepa21} and VideoMAE~\citep{videomae} use 32-frame clips at RGB stride two, spanning 63 source positions. Resizing preserves aspect ratio with short side 256 and dimensions divisible by 16 ($448\times256$ without cropping). A sampled spatial crop is shared across the clip, and both models use ImageNet normalization. Generated clips are decoded in overlapping four-latent pieces with overlap two.

Video-only variants use five projected spatiotemporal heads. The mixed DINOv2+VideoMAE configuration supplements the image heads with a final-layer video branch consisting of spatial pooling within each tubelet, a $768\rightarrow512$ projection, and temporal convolutions that produce scalar logits. Separate heads and optimizers preserve the distinct token organizations of the image and video branches.

The video-only discriminator retains dense spatiotemporal tokens, whereas the auxiliary temporal branch in the mixed configuration pools spatial locations before temporal discrimination. Image encoders process sampled frames independently, while video encoders capture cross-frame variation within the sampled clip. The two branches therefore provide complementary spatial and temporal feature structures.

\textbf{Diffusion-internal representation.}
DiT-GAN replaces the external VFM with frozen teacher features. Real and generated latents receive a shared diffusion timestep and noise realization, with null text conditioning used for feature extraction. Projected features are passed to local and video-pooled heads. Generator updates differentiate through the frozen feature model, while discriminator updates use detached latents.

\textbf{Multi-level sensitivity.}
A local expansion characterizes the sensitivity induced by multiple feature levels. Let $f_\ell(x)$ be a vectorized frozen feature and $a_\ell\geq0$ fixed weights. At a differentiable input, a small perturbation $\delta x$ gives
\begin{align}
 \sum_\ell a_\ell
 \|f_\ell(x+\delta x)-f_\ell(x)\|_2^2
 &=\delta x^{\mathsf T}M_\Phi(x)\delta x
   +o(\|\delta x\|_2^2),\\
 M_\Phi(x)&=\sum_\ell a_\ell J_{f_\ell}(x)^{\mathsf T}J_{f_\ell}(x)
 \succeq0 .
 \label{eq:rad_local_geometry}
\end{align}
For positive weights, the null space of $M_\Phi(x)$ is the intersection of the null spaces of the selected feature Jacobians. Adding feature levels therefore expands the set of locally observable directions whenever the corresponding sensitivities are complementary, while redundant levels contribute less additional information.

Specifically,
$\delta x^{\mathsf T}M_\Phi\delta x
=\sum_\ell a_\ell\|J_{f_\ell}\delta x\|_2^2$
vanishes exactly when every positively weighted feature has zero first-order response. Adding a feature level further intersects this set with the kernel of its Jacobian, reducing or preserving the set of locally invisible directions. The effectiveness of the resulting multi-level discriminator is then determined jointly by feature sensitivity, discriminator heads, optimization, and the generated sample distribution.

\textbf{Encoder size and feature depth.}
The size ablation compares DINOv2-S/B/L/G, all evaluated on the same 15 seeds. The depth ablation keeps DINOv2-S fixed and varies the transformer blocks listed in Table~\ref{tab:rad_encoder_layers}. The layer count excludes the additional patch-input branch. These configurations expose different feature depths while preserving the generator inference architecture.

\begin{table}[t]
\centering
\small
\caption{DINOv2-S feature-depth ablation. All variants retain a patch-input head and use the same 15 evaluation seeds.}
\label{tab:rad_encoder_layers}
\setlength{\tabcolsep}{10pt}
\begin{tabular}{ccc}
\toprule
Transformer layers & Block indices & Total heads\\
\midrule
1 & $\{11\}$ & 2\\
2 & $\{6,11\}$ & 3\\
4 & $\{6,8,10,11\}$ & 5\\
8 & $\{2,4,6,7,8,9,10,11\}$ & 9\\
\bottomrule
\end{tabular}
\end{table}

\section{Additional Visualizations and Prompt Details}
\label{sec:more_visualization_results}

\subsection{Additional Qualitative Comparisons}
\label{app:additional_qualitative}

We provide additional qualitative comparisons across generation regimes and discriminator configurations, complementing the quantitative results in the main text. The cases cover human activities, stylized scenes, animals, and object interactions, including several detailed prompts that specify both appearance and action. Each 5-second case shows three frames at $0$, $2.5$, and $5$ seconds, while each long-video case shows six frames spanning approximately one minute. Prompts and displayed times are shared across methods within each case. 

\begin{figure*}[h]
\centering
\includegraphics[width=\textwidth]{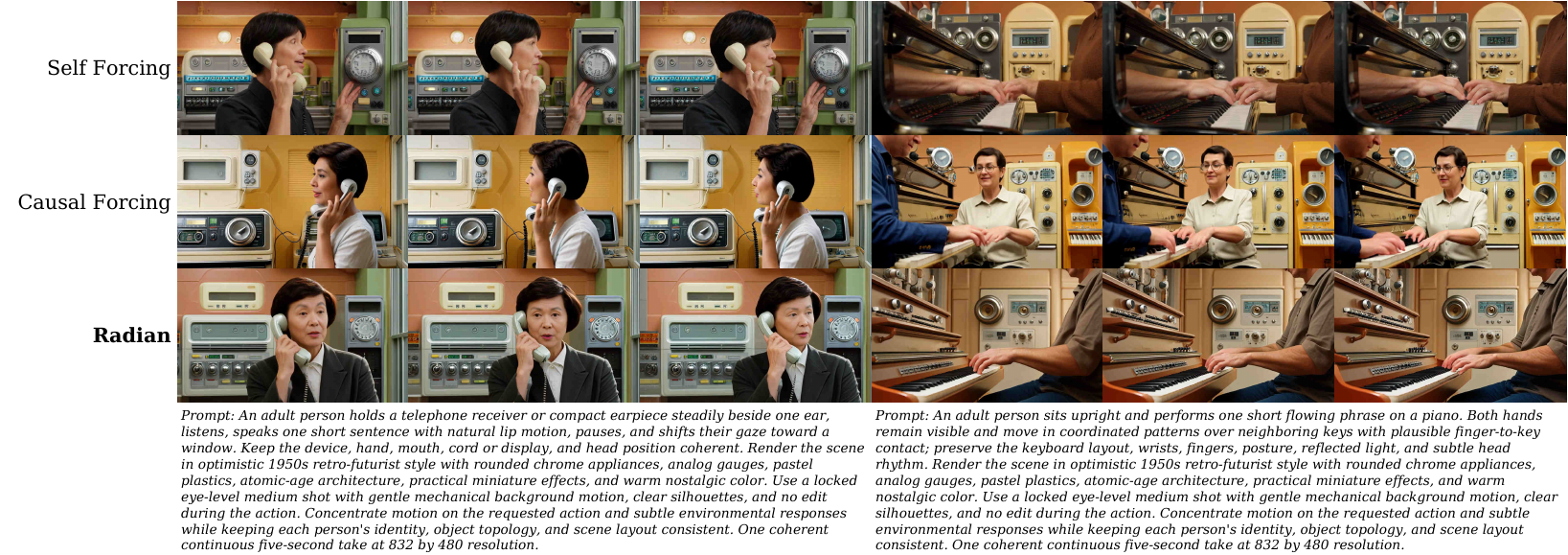}
\caption{\textbf{Additional four-step chunk-wise comparisons.} A retro-futuristic telephone conversation and piano performance, with Self Forcing, Causal Forcing, and Radian arranged from top to bottom. The full prompts appear below the corresponding cases.}
\label{fig:more_chunk_wise}
\end{figure*}

\begin{figure*}[ht]
\centering
\includegraphics[width=\textwidth]{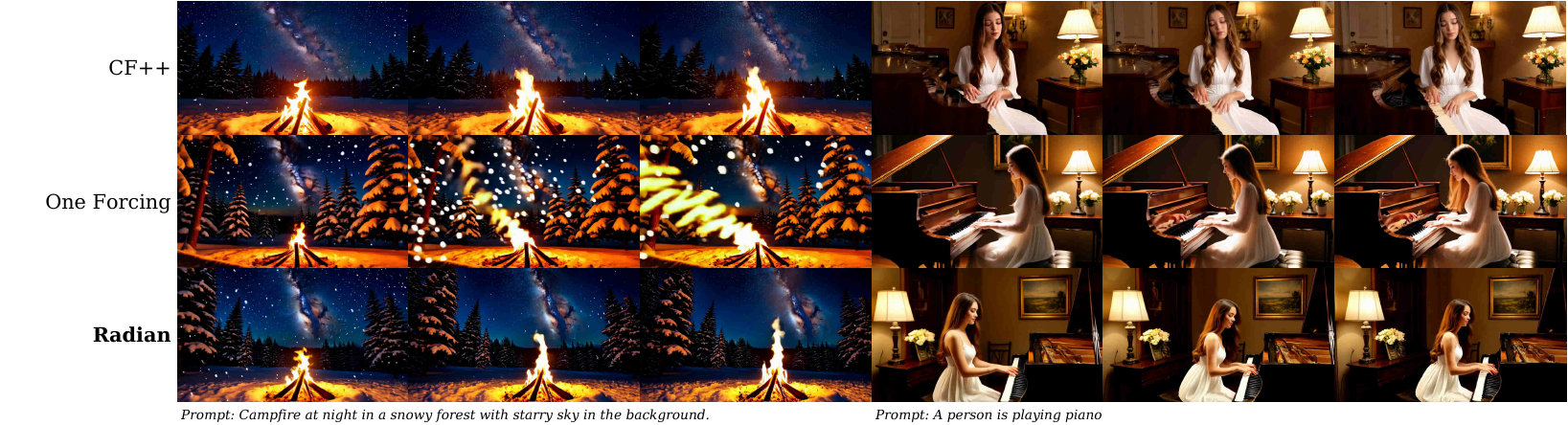}
\caption{\textbf{Additional one-step frame-wise comparisons.} A snowy campfire and piano performance, with Causal Forcing++, One Forcing, and Radian arranged from top to bottom. The sampled frames show changes in scene appearance and subject framing.}
\label{fig:more_frame_wise}
\end{figure*}

\noindent\textbf{Short-video generation.} Figures~\ref{fig:more_chunk_wise} and~\ref{fig:more_frame_wise} compare Radian with Self Forcing~\citep{self_forcing} and Causal Forcing~\citep{causal_forcing} under four-step chunk-wise generation, and with Causal Forcing++~\citep{causal_forcing_pp} and One Forcing~\citep{one_forcing} under one-step frame-wise generation. In the four-step examples, Radian maintains recognizable faces and keeps the telephone receiver or the pianist's hands and keyboard visible across the sampled frames. In the one-step examples, the Causal Forcing++ piano sample does not clearly show the requested keyboard interaction, while One Forcing exhibits larger changes in the apparent scale of the campfire and surrounding trees. Radian retains the piano-playing composition and the broader snowy scene. These observations complement the main-text results on the value of decoded-output supervision for preserving visual structure under few-step generation.

\begin{figure*}[ht]
\centering
\includegraphics[width=\textwidth]{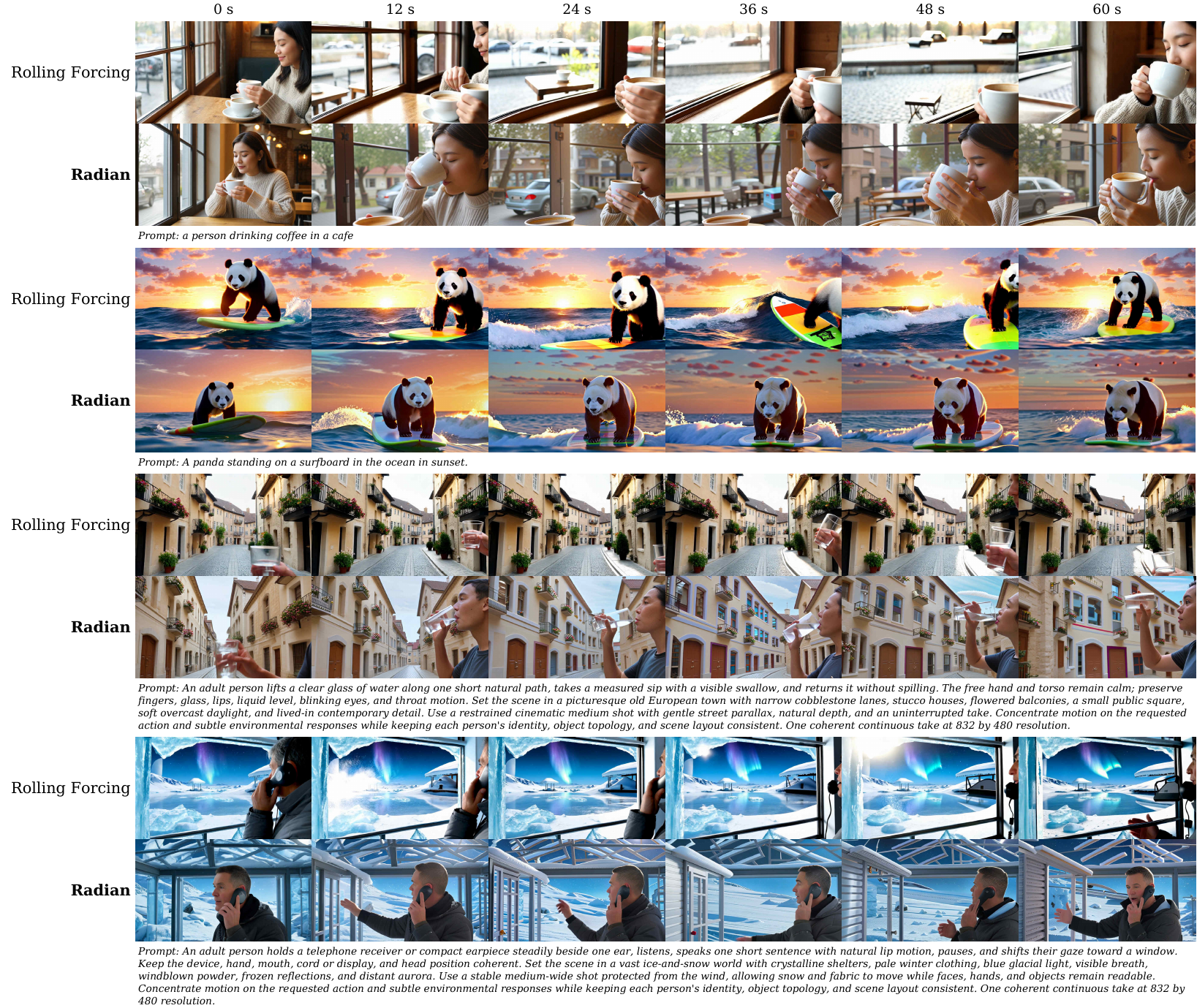}
\caption{\textbf{Additional long-video comparisons.} From top to bottom: coffee in a cafe, panda surfing, drinking water in a European town, and a telephone conversation in the snow. Each case places Rolling Forcing above Radian and its prompt below. Six frames span each 957-frame sequence; the shared time labels are rounded to the nearest second.}
\label{fig:more_long_video}
\end{figure*}

\noindent\textbf{Long-video generation.} Figure~\ref{fig:more_long_video} compares Radian with Rolling Forcing~\citep{rolling_forcing} on four approximately 60-second videos. In the surfing and telephone examples, Rolling Forcing shows larger subject displacements and partial cropping at several sampled times. Radian keeps the panda and surfboard recognizable and retains a clearer view of the person holding the telephone throughout the displayed sequence. The cafe and European-town examples further show changes in viewpoint while preserving identifiable people, objects, and scene context. These examples are consistent with the main-text long-video results, illustrating how the benefits of representation-space supervision can persist beyond the short generation horizon.

\begin{figure*}[ht]
\centering
\includegraphics[width=\textwidth]{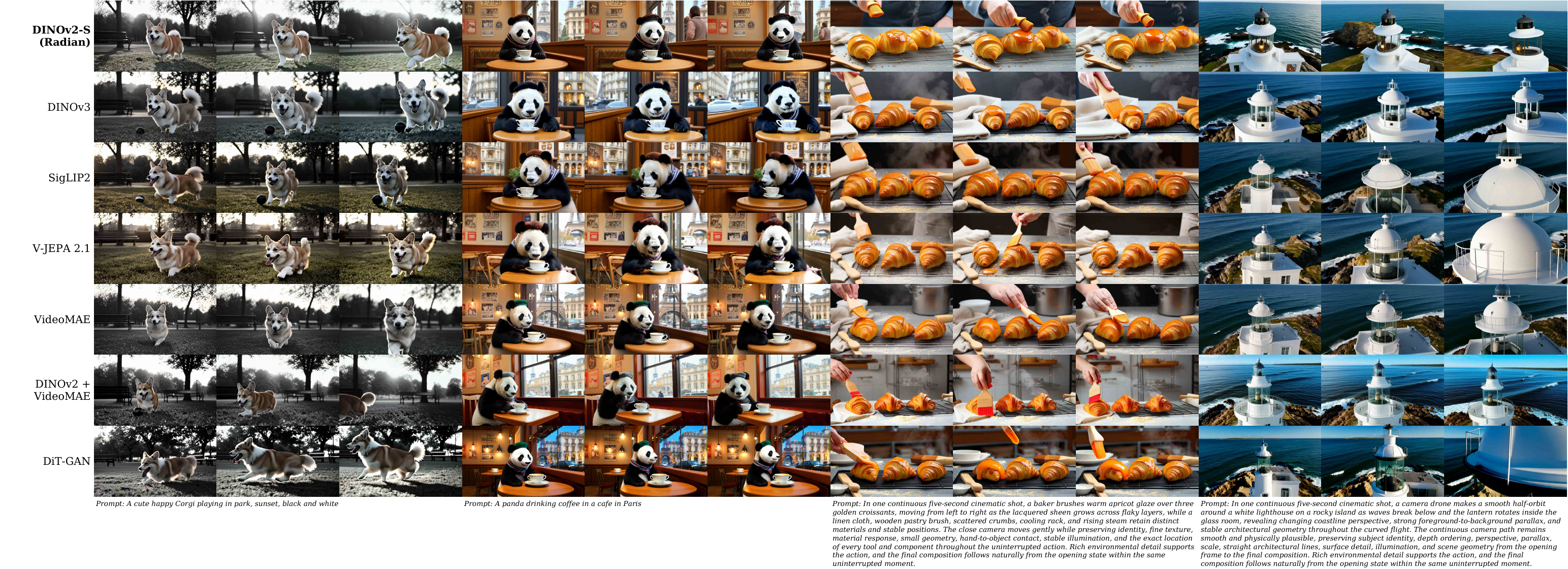}
\caption{\textbf{Feature-space comparisons.} Rows show DINOv2-S, DINOv3, SigLIP2, V-JEPA 2.1, VideoMAE, DINOv2 with VideoMAE, and DiT-GAN. The four cases depict a black-and-white corgi scene, a panda drinking coffee, glazing croissants, and a coastal lighthouse.}
\label{fig:more_feature_ablation}
\end{figure*}

\begin{figure*}[ht]
\centering
\includegraphics[width=\textwidth]{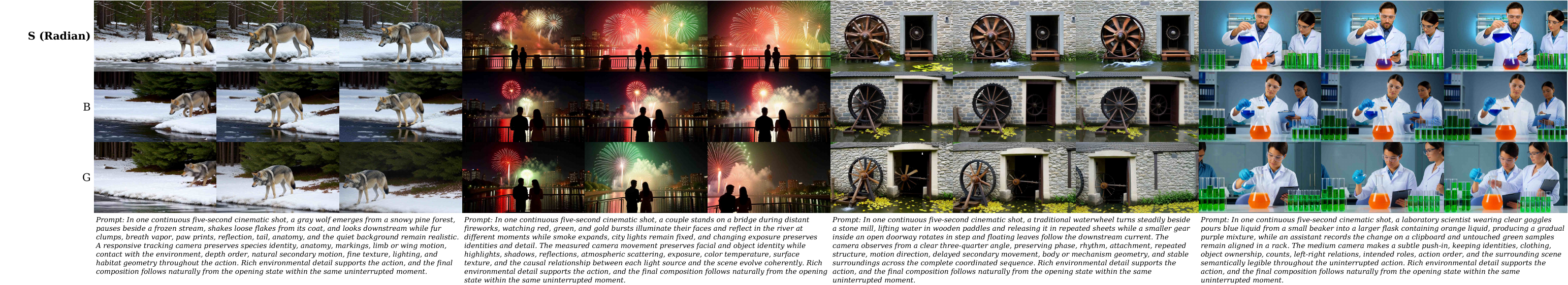}
\caption{\textbf{Encoder-size comparisons.} DINOv2-S, B, and G are shown on a wolf beside a frozen stream, fireworks viewed from a bridge, a traditional waterwheel, and mixing colored liquids. Each case displays three frames with its prompt below.}
\label{fig:more_encoder_size}
\end{figure*}

\begin{figure*}[ht]
\centering
\includegraphics[width=\textwidth]{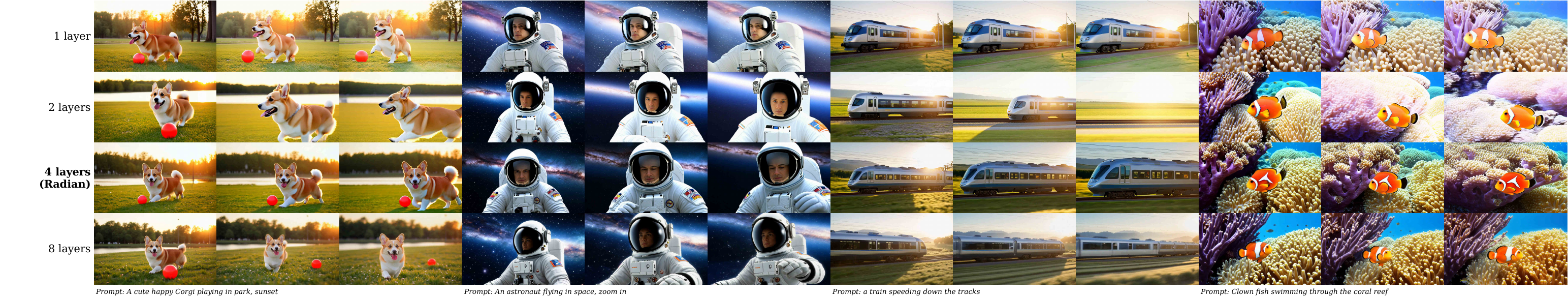}
\caption{\textbf{Feature-layer comparisons.} Rows use one, two, four, and eight DINOv2-S feature layers, respectively. The cases show a corgi in a park, an astronaut, a train on tracks, and clownfish among coral; Radian uses the four-layer configuration.}
\label{fig:more_layer_ablation}
\end{figure*}

\noindent\textbf{Discriminator ablations.} Figures~\ref{fig:more_feature_ablation}--\ref{fig:more_layer_ablation} visualize the feature-space, encoder-size, and feature-layer configurations studied in our ablations. The feature-space examples reveal distinct appearance preferences: VideoMAE follows the black-and-white instruction more closely in the corgi scene, while DINOv2-S retains some color. The coffee, pastry, and lighthouse cases also show differences in object interaction, surface detail, and framing. In the encoder-size examples, the S configuration makes the liquid color transition more apparent, whereas increasing encoder size does not consistently improve the depicted action. The four-layer configuration also combines recognizable structure with local detail in the corgi and astronaut examples. Together, these visualizations complement the quantitative finding that feature suitability matters more than encoder capacity alone, supporting our compact, multi-level DINOv2-S configuration.

\subsection{Prompts for Main-Text Visualizations}
\label{app:main_visual_prompts}

We list the input prompts for the qualitative examples in Figures~\ref{fig:qualitative_41step} and~\ref{fig:qualitative_long}. Short-video prompts follow the left-to-right case order within each generation regime. Long-video prompts follow the upper-left, upper-right, lower-left, and lower-right case order within each method block. Prompt wording is reproduced from the generation records.

\textbf{Four-step chunk-wise generation.}
\begin{enumerate}
\item \textbf{Panda in a cafe.} \emph{A panda drinking coffee in a cafe in Paris, tilt up. }
\item \textbf{Steam train.} \emph{A steam train moving on a mountainside. }
\item \textbf{Boat on the Seine.} \emph{A boat sailing leisurely along the Seine River with the Eiffel Tower in background, pan left. }
\item \textbf{Bicycle.} \emph{a bicycle accelerating to gain speed. }
\end{enumerate}

\textbf{One-step frame-wise generation.}
\begin{enumerate}
\item \textbf{Van Gogh-style riverboat.} \emph{A boat sailing leisurely along the Seine River with the Eiffel Tower in background, Van Gogh style. }
\item \textbf{Panda dining.} \emph{A cute fluffy panda eating Chinese food in a restaurant. }
\item \textbf{Dog drinking water.} \emph{a dog drinking water. }
\item \textbf{Koala playing piano.} \emph{A koala bear playing piano in the forest.}
\end{enumerate}

\textbf{Long-video generation.}
\begin{enumerate}
\item \textbf{Oil-painted shark.} \emph{a shark is swimming in the ocean, oil painting. }
\item \textbf{Couple in the rain.} \emph{A couple in formal evening wear going home get caught in a heavy downpour with umbrellas, Van Gogh style. }
\item \textbf{Trumpet performance.} \emph{A person is playing trumpet. }
\item \textbf{Cooking in a medieval castle.} \emph{An adult person prepares vegetables in a wok, holding the handle firmly with one hand while moving a spatula through two compact circular strokes and one small controlled toss. Ingredients remain inside the pan; preserve both hands, wok, stove, vegetables, steam, and heat source. Set the scene inside a historically textured medieval castle with rough stone walls, carved oak furniture, iron fixtures, wool and linen garments, torchlight, banners, and cool daylight from arrow-slit windows. Use a stable period-film composition that keeps the complete action visible; no cuts, whip pans, or sudden costume transformations. Concentrate motion on the requested action and subtle environmental responses while keeping each person's identity, object topology, and scene layout consistent. }
\end{enumerate}


\end{document}